\documentclass[iicol]{sn-jnl}

\usepackage{cite}
\usepackage{amsmath,amssymb,amsfonts}
\usepackage{graphicx}
\usepackage{textcomp}
\usepackage{xcolor}
\usepackage{caption}
\usepackage{subcaption}
\usepackage{adjustbox}
\usepackage{array}

\newcommand{\ndedit}[1]{\textcolor{black}{#1}}
\newcommand{\ndedittwo}[1]{\textcolor{black}{#1}}

\let\orgautoref\autoref
\providecommand{\Autoref}[1]
{%
\def\equationautorefname{Equation}%
\def\figureautorefname{Figure}%
\def\subfigureautorefname{Figure}%
\def\algorithmautorefname{Algorithm}%
\def\sectionautorefname{Section}%
\orgautoref{#1}%
}

\jyear{2023}%

\theoremstyle{thmstyleone}%
\theoremstyle{thmstyletwo}%

\theoremstyle{thmstylethree}%

\begin{document}

\title[Article Title]{\centering Designing Robot Identity: \\The Role of Voice, Clothing, and Task\\ on Robot Gender Perception}


\author*[]{\fnm{Nathaniel} \sur{Dennler}}\email{dennler@usc.edu}

\author[]{\fnm{Mina} \sur{Kian}}\email{kian@usc.edu}
\author[]{\fnm{Stefanos} \sur{Nikolaidis}}\email{nikolaid@usc.edu}
\author[]{\fnm{Maja} \sur{Matari\'c}}\email{mataric@usc.edu}

\affil[]{\orgdiv{Computer Science Department}, \orgname{University of Southern California}, \orgaddress{\city{Los Angeles}, \state{CA}, \country{USA}}}


\abstract{

 Perceptions of gender have a significant impact on human-human interaction, and gender has wide-reaching social implications for robots intended to interact with humans. This work explored two flexible modalities for communicating gender in robots--voice and appearance--and we studied their individual and combined influences on a robot's perceived gender. We evaluated the perception of a robot's gender through three \ndedit{online studies}. First, we conducted a \textit{\ndedit{voice design} study} (n=65) on the gender perception of robot voices by varying speaker identity and pitch. Second, we conducted a \textit{\ndedit{clothing design} study} (n=93) on the gender perception of robot clothing designed for two different tasks. Finally, building on the results of the first two studies, we completed a large \textit{integrative video study} (n=273) involving two human-robot interaction tasks. We found that voice and clothing can be used to reliably establish a robot's perceived gender, and that combining these two modalities can have different effects on the robot's perceived gender. Taken together, these results inform the design of robot voices and clothing as individual and interacting components in the perceptions of robot gender.
}




\maketitle

\section{Introduction}

Robots are increasingly moving from being purely \textit{functional} devices that work in isolated contexts to taking on \textit{social} roles that engage with, support, and interact with humans~\cite{pandey2018mass,specian2021quori,suguitan2019blossom}. This shift of contexts introduces new design considerations about how a robot's social identity affects its use, in addition to its functional role~\cite{deng2018formalizing}.

Several theoretical underpinnings support the idea that technology has unavoidable and salient social implications. The paradigm of computers as social actors \cite{nass1994computers} has accumulated a long-standing and in-depth body of work showing that computers that are deployed in social contexts are viewed as social agents, regardless of whether or not they have social agency \cite{nass2000machines}. In the case of robots, this effect is further reinforced by the robot's physical embodiment \cite{wainer2006role,deng2019embodiment}. \ndedit{Embodiment provides additional interaction modalities--such as gesture, gaze, and movement--that reinforce social identity. Understanding how robots may use these additional modalities to establish identity is crucial for robot deployments in human-facing tasks because ecological use is affected by both user and robot  identity \cite{esmaeilzadeh2021does,devito2018too,tapus2008user}.}

Beyond use considerations, identity affects how well people can complete collaborative tasks. Findings from social psychology, economics, operations research, and marketing research show that social identity strongly influences both subjective and objective interaction metrics \cite{tajfel2004social, white2019shift, hennessy1999intergroup,charness2020social}. \ndedit{The underlying process that drives changes in these metrics is the formation of different social groups, as described by Social Identity Theory \cite{tajfel2004social,stets2000identity}.  Social Identity Theory describes how people in \textbf{the same} social identity groups attribute more \textbf{positive} characteristics and take more \textbf{favorable} actions toward other members of their social identity, regardless of whether the in-group member is a human or a robot \cite{fraune2020our, fraune2017teammates, haring2014would,kuchenbrandt2013robot,sebo2020robots}. Conversely, members of \textbf{different} social identity groups typically behave more \textbf{negatively} toward each other \cite{davis2019collective}; this effect also applies to both human-human and human-robot interaction \cite{sebo2020robots, chang2012effect,fraune2019human}. For example, Fraune et al. \cite{fraune2017teammates} explored the effect of Social Identity Theory in a study with two mixed teams of two humans and two robots each competed in a price-is-right game. They found that participants subjectively rated the in-group robot as more cooperative than out-group humans, and participants assigned more painful noise blasts to out-group humans than to in-group robots. While group structures in game contexts are well-defined, groups formed in daily life are more contextual. However, similar effects are well established in real-world human groups based on identity traits such as gender \cite{charness2020social}, political affiliation \cite{hart2012boomerang}, and even dietary habits \cite{davis2019collective}. To understand how a robot's identity affects use and adoption, designers must accurately understand how the robot's identity is perceived.} 

\ndedit{In this this work, we focus on a particular aspect of a robot's identity: \textit{gender}.} {\it Gender} is one of the most widespread modalities of social identity, and has been linked to several interaction differences in computer interfaces \cite{bardzell2010feminist,stumpf2020gender} as well as in robots \cite{kuchenbrandt2014keep,sandygulova2018age}. \ndedit{Gender is a highly contextual and complicated form of identity,} however, previous works on gender perception in robotics have \ndedit{adopted simplistic frameworks to investigate gender that} only consider a single sensory modality to modify the robot's perceived gender \cite{steinhaeusser2021anthropomorphize, law2021touching, crowelly2009gendered, raghunath2022robot, raghunath2021women}. \ndedit{Those works assumed that single-modality effects can be linearly combined to understand the impact that a robot's gender has on interaction. In this work, we show that is not the case. While some} past research has considered multiple modalities to \ndedit{gender a robot, it generally explored normative views of gender and designed the robot to be} wholly masculine or wholly feminine \cite{chita2019gender, kuchenbrandt2014keep, powers2005eliciting,bryant2020should, tay2014stereotypes, eyssel2012s}. In contrast, research in philosophy, feminism, queer theory, and Human-computer interaction (HCI) has found that gender is highly nuanced and formed through a complex interaction of several modalities and social interactions \cite{butler2002gender}. The current simplifications present in the robotics formulations of how robots are gendered may lead to the inconsistencies present in the literature, such as finding the influence of robot gender on a user's interaction with the robot as significant in some cases \cite{chita2019gender,crowelly2009gendered,eyssel2012s,tay2014stereotypes,kuchenbrandt2014keep,powers2005eliciting,raghunath2022robot}, and having no effect in others \cite{law2021touching,steinhaeusser2021anthropomorphize,bryant2020should,robben2023effect, raghunath2021women,paetzel2016effects}, \ndedit{discussed in detail throughout \Autoref{related}}. By understanding how different sensory modalities interact, we can more effectively interpret those inconsistencies and further the field's understanding of those complex influences at a time when robots are actively being developed for use in human daily lives.

\ndedit{To expand the understanding of how gender is attributed to robots in human-robot interaction (HRI) studies from a queer and feminist perspective, we contribute the following: (1) a framework to evaluate the perceived gender characteristics of a robot's voice, (2) a design methodology to develop clothing for robots, and (3) a quantitative evaluation showing that gender perception is not a linear combination of its constituent parts. }

We posit that the construction of a robot's gender is an important part of the design process that needs to be considered for each context. To address this challenge, we outline a design process and design principles for eliciting a particular perception of the robot's gender in robots through voice and appearance.  Based on those principles, we designed and conducted three user studies that explored users' perceptions of a robot's voice and appearance, separately and then together, in two tasks with different social roles, reinforcing some expected stereotypes and uncovering some novel insights. We found that the perception of a robot's gender can be modulated through the careful design and evaluation of both modalities and that the perceived gender of a robot is influenced by the robot's task. We present new results about the construction of robot gender, and the relative influences of voice, appearance, and task.  


This paper is organized as follows. \Autoref{related} reviews the related work in gender construction and gender in human-machine interaction. \Autoref{voice_study} describes our \ndedit{voice design study that we used to design} a voice that can vary gender perception. \Autoref{clothing_study} describes our \ndedit{clothing design study that we used to design clothes that} elicit different gender expressions for two tasks: a medical professional and a hotel receptionist. \Autoref{video_study} describes our integrative video study that combines the results from the first two studies to investigate how auditory and visual modalities interact with tasks of different social roles to establish a robot's perceived gender. \Autoref{discussion} discusses the results to generate insights from the three studies and concludes the paper.

\section{Background}\label{related}
Our work applies insights from philosophy, feminism, \ndedit{queer theory}, and HCI to the domain of robotic design. In the following section we describe these insights and review how gender has previously been explored in the context of robotics.

\subsection{\ndedit{Feminist and Queer Conceptualizations of Gender}}

\ndedit{Historically, researchers across many fields have conflated two fundamentally distinct constructs: \textit{sex} (e.g., male or female), a person's biological category related to their genetic makeup, and \textit{gender} (e.g., man or woman), a person's social category related to their behaviors in society \cite{west1987doing}. Previous longstanding and problematic conceptualizations of gender described gender as immutable, binary, and physiological \cite{keyes2018misgendering}, where \textit{binary} refers to the presence of only two labels (man and woman), \textit{immutable} refers to the inability to change the label ``man" or ``woman" once it has been established, and \textit{physiological} refers to gender being assigned based on physically expressed characteristics of a person.}

\ndedit{Modern feminist and queer researchers have identified that none of these previously adopted properties of gender describe how societies actually perceive gender \cite{messerschmidt2009doing}. For example, trans people are people whose gender does not align with their sex assigned at birth \cite{scheurmangender2020}. The existence of trans people directly conflicts with the idea of immutable gender. Non-binary people do not fit neatly into the labels ``man" or ``woman" \cite{mcnabb2017nonbinary}. The existence of non-binary people conflicts with the idea of a binary gender. Intersex people are born with physiological characteristics that do not directly match the criteria for either sex \cite{preves2003intersex}, and ethnographic researchers have found that people make determinations of gender with non-physiological cues, such as posture, dress, or vocal cues \cite{kessler1985gender}. Both of these ideas conflict with the notion of physiological gender. Previously held beliefs of gender have excluded queer and trans people from research \cite{namaste2000invisible,queerinai2023queer}, and have influenced the way that gender is perceived in robotics research. We highlight two important ideas that were created by queer and feminist scholars to examine how these misconceptions have impacted robotics research: \textit{feminist standpoint theory}, and \textit{gender performativity}. Precise language is critical for discussing nuanced topics; thus we adopt the recommended terminology from the HCI gender guidelines \cite{scheurmangender2020}: ``male" and ``female" are only used to refer to sex, ``men" and ``women" are used to refer to people, and ``masculine" and ``feminine" are used to refer to items that may be associated with 
a gender but do not intrinsically have a gender. }

\noindent\textbf{Feminist Standpoint Theory.}
\ndedit{\textit{Feminist standpoint theory} is an epistemology developed by feminists to describe how the construction of knowledge is affected by power structures. Feminist standpoint theory has been used to develop several research agendas, across HCI \cite{bardzell2010feminist}, HRI \cite{winkle2023feminist}, and the social sciences \cite{rayaprol2016feminist}. This theory provides four guiding theses described in detail by Gurung \cite{gurung2020feminist}, summarized here: (1) strong objectivity requires marginalized perspectives--what is described as a fact must be agreed upon from people of multiple perspectives to be true, (2) social context (i.e., a person's \textit{standpoint}) shapes and limits what can be learned--members of a socially advantaged group may not be aware of the experiences of other marginalized groups and cannot perform research that generates the knowledge held by these marginalized groups, (3) marginalized people are acutely aware of their experience--many rules and regulations are created by the socially advantaged members of society, so marginalized populations are more likely to be aware of how these rules and regulations oppose their own experience, and (4) power dynamics distort evidence--researchers that are not part of a marginalized group may not be able to collect holistic data from a marginalized group due to historical mistreatment of the group.}

\ndedit{In this work, we investigate how a robot's gender is constructed from the perspective of feminist standpoint theory. First, we present the creation of a robot's identity as a design problem. This allows multiple perspectives to be included in the construction of a gendered robot, thereby providing an avenue for marginalized identities to incorporate their knowledge into the development of robots. Second, we incorporate queer and non-binary perspectives into that design process. Queer and non-binary people are marginalized groups that possess knowledge on the construction of gender, yet are often underrepresented in robotics and AI research \cite{korpan2024launching,queerinai2023queer}.}

\noindent\textbf{Gender Performativity.}
\ndedit{The third-wave feminist movement emphasized the idea of gender as being \textit{socially constructed}, rather than an objective fact. In particular, Butler reconceptualized gender as being \textit{performative}--i.e., defined by a sequence of acts that reinforce a particular identity \cite{salih2007judith, butler2002gender,west1987doing}. This means that a person's gender is created by repeatedly performing actions that align with society's perception of a gendered role. Concretely, a person is a woman because they repeatedly perform feminine actions expected of a woman, not because they are born a woman. This modern view of gender more closely aligns with the way people interpret gender and refutes the previous perspective of a binary, immutable, and physiological gender.}

\ndedit{Core to the conceptualization of a performative gender is the idea of choice. In robotics, designers make the choices about a robot's appearance and behaviors, yet users typically perceive a robot as being agentic \cite{jackson2021theory}. Therefore, choices made by designers may instead be attributed to robots instead, establishing a robot's identity beyond what the designer originally intended. In this work, we investigate how design choices made by robot designers can be interpreted as intentional choices by the robot to communicate a particular gender. This stands in contrast to the prior assumption that robots are designed to be a particular gender and people assign robots genders from that design \cite{nomura2017robots}.}



\subsection{\ndedit{Ethical Considerations for} Designing Robot Gender}

\ndedit{Designing any artifact, physically embodied or not, is an inherently political activity because the creators of artifacts imbue those artifacts with their personal values and preferences \cite{friedman1996value,winner2017artifacts}. The values imbued in the design of robots have the potential to negatively impact the lives of marginalized communities by perpetuating or enforcing harmful stereotypes about these communities. While this work specifically addresses marginalized genders, these harms extend beyond gender considerations. Importantly, people who have multiple intersecting marginalized identities are especially susceptible to harms caused by not being involved in the design process \cite{costanza2018design}.}

\ndedit{Harms to marginalized communities have been widely observed in several related fields. For example, UNESCO released a report on voice assistants that describes how a lack of women designing voice assistants has led to voice assistants reflecting several problematic stereotypes of women. Feminine voice assistants are designed to be obliging and eager to please while also being tolerant of abuse from users \cite{west2019d}. Portraying voice assistants in this stereotypical way can directly lead to the unfair treatment of women in general \cite{weisman2020instantaneity}. Other AI systems also lack diverse developers and thus perpetuate harmful stereotypes of marginalized populations especially due to the biased design of the datasets these systems are trained on \cite{birhane2021multimodal,birhane2021large,scheuerman2021datasets}. Large Language Models (LLMs) have been shown to generate text that perpetuates gender biases~\cite{wan2023kelly}, homophobia~\cite{sheng2019woman}, transphobia~\cite{dev2021harms}, ableism~\cite{azeem2024llm}, racism~\cite{lee2019exploring}, hate speech~\cite{hartvigsen2022toxigen}, abusive langauge~\cite{waseem2017understanding}, dehumanization~\cite{mendelsohn2020framework}, and microaggressions~\cite{gross2023chatgpt} against marginalized populations. In addition to LLMs, other perceptual modules that are designed without marginalized communities have historically excluded nonbinary people~\cite{keyes2018misgendering,scheuerman2019computers}, failed at higher rates for intersecting marginalized identites like race and gender~\cite{buolamwini2018gender}, and labeled objects with offensive language and slurs~\cite{birhane2021large}. }

\ndedit{Robots combine several AI systems with physical affordances that can harm marginalized groups and thus can inherit and exacerbate these stereotypes and biases \cite{hundt2022robots,azeem2024llm}. Robots also have unique considerations due to their physical embodiment \cite{deng2019embodiment}.} For example, robot embodiments tend to be perceived as masculine by default \cite{perugia2022shape}, and gendered robots are often subjected to similar stereotypes as people \cite{perugia2023models} because people routinely assign social traits to interactive systems~\cite{nass1994computers} and automatically place such systems into social categories \cite{nass2000machines}. In the case of robots, their physical embodiment amplifies the degree to which social traits are assigned~\cite{deng2019embodiment, dennler2022using}. 

\ndedit{To address the harms presented by robotic systems being designed without the participation of marginalized populations, researchers have proposed several ethical design frameworks. Many of those frameworks build on the ideas of the principles of \textit{Participatory Design} \cite{spinuzzi2005methodology} and \textit{Design Justice} \cite{costanza2020design}, which emphasize the importance of understanding the power dynamics between designers and users, and call for designers to consider ways in which marginalized users can provide feedback to the design process. For example, the \textit{Feminist HRI Framework} proposed by Winkle et al. \cite{winkle2023feminist} describes feminist principles for conducting HRI research through examining how robots are positioned in an interaction, how users' experiences have shaped their interpretation of the robot, and understanding how research can challenge existing power structures in society. The \textit{HRI Equitable Design Framework} proposed by Ostrowski et al. \cite{ostrowski2022} expands the seven areas proposed by the Design Justice Framework with six more areas that apply to robotics research: entry and exit, autonomy, transparency, deception, futures, and histories. The \textit{Robots for Social Justice} framework proposed by Zhu et al. \cite{zhu2024robots} describes techniques that designers can employ to incorporate feedback from marginalized communities: listening contextually, identifying structural conditions, acknowledging political agency, increasing opportunities, and reducing imposed risks and harms. }

\ndedit{These frameworks provide guiding principles to help designers engage with marginalized communities, and are designed to be broad. In this work, we develop a specific design process that can be readily deployed with the communities that interact with robots, and can be readily used by robot designers to incorporate user feedback. This design process works to subvert the power dynamic between the designer and the user by allowing the user to evaluate and modify design decisions with respect to how the robot's gender is perceived.}

\renewcommand{\arraystretch}{1.5}
\newcolumntype{M}[1]{>{\centering\arraybackslash}m{#1}}
\newcolumntype{R}[1]{>{\raggedleft\arraybackslash}m{#1}}

\begin{table*}
    \centering
    \caption{\ndedit{Summary of robot gender manipulation studies. Construct refers to the construct that was evaluated, robot denotes the robot used in the study, gender manipulation denotes what aspect of the robot was changed, robot task describes what participants viewed the robot doing, and finding summarizes the effect that a gendered robot on participant perception. All gender manipulations with a ``+" refers to changing all listed modalities at the same time to construct the robot's gender, whereas a ``," denotes independent manipulation. Robot tasks with a ``," denote different experimental conditions.}}
    \adjustbox{max width=\textwidth}{%
    \begin{tabular}{R{5cm} M{1.7cm} M{2.1cm} M{3cm} M{8cm}}
        \toprule
        \textbf{Construct} & \textbf{Robot} & \textbf{Gender Manipulation} & \textbf{Robot Task} & \textbf{Finding} \\
        \midrule

        Duration of Task~\cite{kuchenbrandt2014keep} & NAO & name + voice & sorting tools, sorting sewing equipment & Men completed tasks faster when working with a masculine robot compared to a feminine robot. \\
        
        Reliability~\cite{crowelly2009gendered} & ActivMedia PeopleBot & voice & survey administration & The masculine robots and feminine speaker were rated as more reliable than masculine speaker and feminine robot. \\
        
        Emotional Intelligence~\cite{chita2019gender} & PR2 & name + voice & workplace discussion & The masculine robot was rated as more emotionally intelligent than the feminine robot for the same dialogue. \\
        
        Acceptance~\cite{tay2014stereotypes} & Olivia & name + voice & healthcare, security & The masculine robot had higher acceptance than the feminine robot for the security task, and lower acceptance than the feminine robot for the healthcare task. \\
        
        Agency, Communion~\cite{eyssel2012s} & Flobi & hair length & N/A (image) & The masculine robot was rated as higher in agency and lower in communion than the feminine robot. \\

        Number of Words Spoken~\cite{powers2005eliciting} & Nursebot & voice + lip color & discuss dating advice & Participants said more words to the masculine robot about dating advice than the feminine robot. \\
        
        \midrule

        Transportation, Anthropomorphism~\cite{steinhaeusser2021anthropomorphize} & NAO & voice & story telling & No gender-based effects on transportation or anthropomorphization. \\

        Trust~\cite{law2021touching} & PR2 & voice & data entry supervision & No gender-based effect on trust. \\
        
        Warmth, Competence, Discomfort, and Joke Rating~\cite{raghunath2022robot} & NAO & voice & comedy performance & No gender-based differences in Competence, Discomfort, or Joke Rating. The feminine robot was rated as more Warm than the masculine speaker. \\

        Warmth, Competence, Discomfort, and Joke Rating~\cite{raghunath2021women} & NAO & voice & comedy performance & No gender-based differences on Warmth, Competence, Discomfort, or Joke Rating. \\
        
        Occupational Competency, Trust~\cite{bryant2020should} & Pepper & name + voice & introducing itself & No gender-based effects on occupational competency or trust. \\
        
        Acceptance, Anthropomorphism, and Enjoyment~\cite{robben2023effect} & NAO & name + voice + accessory & story telling & No gender-based differences in Acceptance, Anthropomorphism, or Enjoyment. \\

        Eeriness~\cite{paetzel2016effects} & Furhat & voice, face & introducing itself & No gender-based differences in Eeriness. \\
        
        \bottomrule
    \end{tabular}%
    }
\end{table*}
\subsection{Gender Manipulations in Robotics \ndedit{Research}}

We briefly describe how past studies have implemented robot manipulations that are claimed to modify the perceived gender of robots, and note that several studies made manipulations without verifying the effect of those manipulations on the process of gendering robots \cite{seaborn2022pronouns}. \ndedit{In this work, we aim to explicitly evaluate how gender is perceived when combining different modalities of interaction.}


\noindent\textbf{Auditory Gender Manipulations.} Due to the widespread deployment of voice assistants \cite{west2019d}, gender in synthesized voices has been studied \cite{nass2005wired,seaborn2021voice}. Pitch was found to be the key component of perceived gender in voice \cite{tolmeijer2021female,seaborn2021voice}, however several studies that manipulated gender chose from pre-made TTS voices with pre-set genders \cite{steinhaeusser2021anthropomorphize,bryant2020should,law2021touching,chita2019gender, raghunath2021women,raghunath2022robot}. A review of voices in human-agent interaction has found that, in addition to gender, voice also establishes other important identity characteristics through other voice properties, including accent, speech style, and loudness \cite{seaborn2021voice}. Additionally, the perceived attributes of a voice depend on the embodiment that generates the voice \cite{mcginn2019can,cambre2019one}. In this work, we make subtle voice manipulations to study the effect of voice pitch on gender, while minimizing changes to other possible identities of the voice.

\noindent\textbf{Visual Gender Manipulations.} Differences in facial cues~\cite{eyssel2012s}, accessories~\cite{jung2016feminizing, neuteboom2021cobbler,robben2023effect}, and body proportions~\cite{trovato2018she, trovato2017influence,bernotat2021fe} have been used to provide different perceptions of robot gender. Due to physical limitations, the actual embodiment of a robot is difficult to change, however clothing has been identified as a way to change perceived robot embodiment \cite{friedman2021robots}. In this work, we extrapolate this idea and use the concept of design elements from fashion, such as silhouette and lines \cite{sorger2017fundamentals}, to establish robot gender identities.

\noindent\textbf{Robot Gender and Task.} In addition to establishing a robot's identity, the perceived gender of robots has been linked to differences in \ndedit{objective measures of} task performance \cite{powers2005eliciting,kuchenbrandt2014keep}. Past work comparing robots in security and healthcare tasks found that participants reported higher acceptance when a robot's gender matched the expected gender for the task \cite{tay2014stereotypes}. However, in another study, participants showed preferences for anthropomorphic robots to be masculine, regardless of the specific task context \cite{roesler2022context}. Similarly, work by Bryant et al. \cite{bryant2020should} found no gender-based differences in occupational competency or trust across 17 different occupations. While the results of how task and gender interact are mixed, we hypothesize that task is important in designing a robot's gender \cite{acskin2023gendered}.

\noindent\textbf{Gender Neutrality in Robots.} One approach to designing robot gender to reduce stereotypes is to design robots that are gender-neutral \cite{tolmeijer2021female}. We make a distinction in this work between androgynous robots and non-gendered robots. Our previous work has found that animal-like robots are more likely to be non-gendered--i.e., unassociated with either masculine or feminine qualities \cite{dennler2022using}. Androgyny or gender-ambiguity instead corresponds with being associated with both masculine and feminine traits \cite{seaborn2022pronouns,bem1981bem,sutton2020gender}. Past work has shown the benefits of gender-ambiguous voices for mitigating gender-based stereotypes \cite{torre2023can}, and gender-ambiguous appearance can mitigate gender-based stereotypes by promoting non-normative design \cite{perugia2021gender} and allowing users to decide the perceived gender to the robot themselves \cite{bradwell2021design}.

\section{Voice Design Study}\label{voice_study}

We performed an initial design study of the perceived genders of various voices in order to inform our manipulated voice for the integrative video study, detailed in Section ~\ref{video_study}.\ndedit{The integrative video study aimed to investigate how robot gender is constructed through voice and clothing in interactions with users in professional settings}. \ndedit{First, we formulate our design goals. Next, we describe the design process for the voices aiming to achieve those goals. Finally, we report on the online user study and the evaluation of how well the stated design goals were achieved.}

\subsection{Design Principles}\label{design_principles}
Based on the results of previous studies in voice perception, we developed the following design principles (DPs) for the robot's voice:
\begin{itemize}
    \item \textbf{DP1: Pitch Modulates Gender Perception} The perceived gender of the voice should change as the fundamental frequency of the voice changes. While several studies have linked voice pitch and perceived gender \cite{pernet2012role, puts2006dominance}, there are other factors that contribute to the assignment of gender in voices \cite{cartei2013effect}. To evaluate the interaction of physical appearance and gender, the voice should be able to alter its perceived gender.
    
    \item \textbf{DP2: Clarity and Realism are Consistent} The voice should be similarly understandable and realistic for all perceived genders and pitch modulations. \ndedit{Clarity and Realism} have been linked to user acceptance in digital assistants \cite{cambre2020choice}, and should be held constant across different perceived genders.
    
    \item \textbf{DP3: Identity Follows Function} The voice should \ndedit{be perceived as} professional \ndedit{across different pitches, as we intend to use this voice in professional contexts in the integrative video study}. While perceived gender is one salient feature of a voice, other \ndedit{important aspects of} identity are also expressed through voice, such as age, personality, and geographic region \cite{cambre2019one}. 
\end{itemize}

\begin{figure*}[ht]
    \centering
    \includegraphics[width=\textwidth]{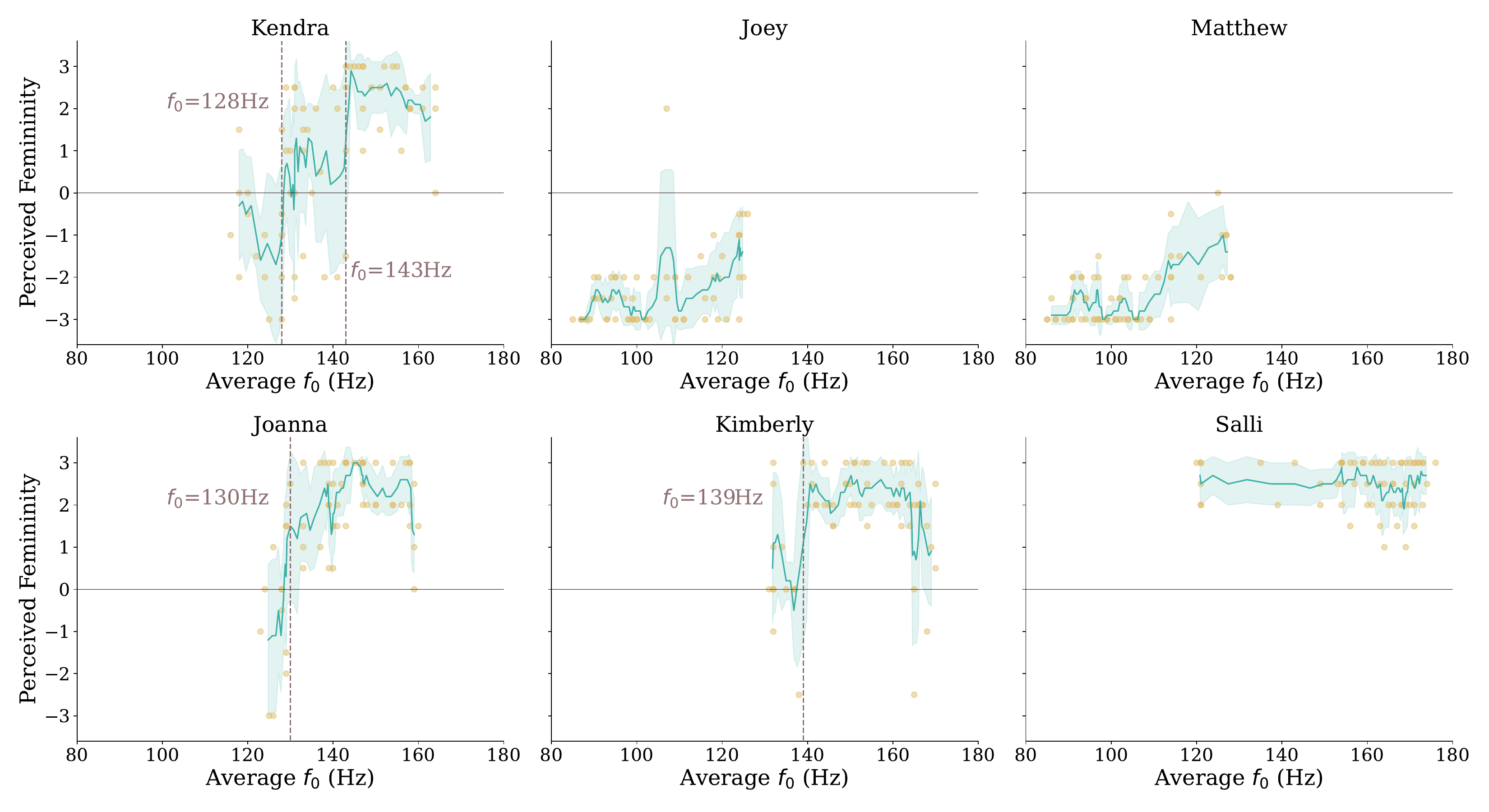}
    \caption{Perceived femininity of voices (-3 represents a masculine voice, 0 represents an ambiguously gendered voice, and +3 represents a feminine voice) as a function of average fundamental frequency ($f_0$) of the utterance. Teal lines represent five-datapoint sliding averages, the shaded region denotes $\pm$ one standard deviation, and the beige dots are individual responses.}
    \label{fig:voice_frequencies}
    \vspace{-0.4cm}
\end{figure*}

\subsection{Design Choices}

\ndedit{For our voice design study, we selected the state-of-the-art text-to-speech service from AWS called} Polly \cite{amazonPolly} to generate synthetic robot voices. \ndedit{Polly has three key benefits. First, it is highly intelligible to users, ensuring that social perceptions of the robot are not tied to a user's difficulty in comprehending the generated speech of the robot. Second, Polly }provides viseme information to automatically synchronize robot mouth movement to spoken words. This establishes the embodiment that produces the speech, which is required for our interaction design. \ndedit{Finally, Polly provides a variety of voice identities, enabling evaluation of multiple options for generating robot speech. At the time of this study, Polly has six adult voices that spoke US English: Joanna, Joey, Kendra, Kimberly, Matthew, and Salli. There were also two child voices, Justin and Ivy. Given that the robot is meant to operate in professional settings, we deemed child-like voices to be inappropriate and selected the six adult voices for evaluation.}

\ndedit{To evaluate DP1, we modified the six existing adult voices to generate additional ones, by changing the pitch by a random value between -3 and +3 semitones using the Python package PyRubberBand \cite{pyrubberband}. A semitone refers to the amount that a sound is shifted in the pitch domain.  For example if our audio was an instrument playing the note $C$, shifting by one semitone resulted in a $C\#$ and two semitones resulted in a $D$. The range of -3 to +3 semitones ensures that the audio remains natural-sounding. We additionally calculated the average fundamental frequency ($f_0$) for all of the modified voices using the librosa Python package \cite{mcfee2015librosa} to obtain a meaningful quantity to compare sounds.}

\subsection{Study Description}

We performed a within-subjects study of the perceived gender of \ndedit {the generated} voices. \ndedit{Our goal was to evaluate each voice in isolation to better understand how it interacts with other features of the robot. To achieve this goal,} each study participant in this study evaluated six voices. \ndedit{The voices were presented as an audio file with no contextual clues that may bias participants' perceptions of gender, i.e., voice names, fundamental frequencies, and intended tasks were all unknown to the participants. We presented the voices in a randomized and counter-balanced order, and each had a random pitch modification within the range of -3 and +3 semitones. Following standard practices, we evaluated each voice by generating sentences that had a neutral sentiment \cite{torre2023can}. We used sentences that described text-to-speech and lasted 12-18 seconds, because previous works that evaluated perception of TTS voice recommend at least 10 seconds of audio \cite{cambre2020choice}. To evaluate the voices,} participants rated the \ndedit{following 7-point Likert items (based on Cambre et al. \cite{cambre2020choice}) ranging from ``Strongly Disagree" to ``Strongly Agree":}
\begin{enumerate}
    \item The voice sounds feminine (feminine)
    \item The voice sounds masculine (masculine)
    \item The voice sounds like a real person (realism)
    \item The voice is easy to understand (clarity)
\end{enumerate}
\ndedit{Participants also} provided answers to an open-ended question detailing their personal perception of each voice \ndedit{by responding to the prompt ``describe how the voice sounds to you in one to two short sentences".} 

\ndedit{We administered this study, approved under USC IRB \#UP-18-00510, using Amazon Mechanical Turk.} The participants first filled out an informed consent form and entered their demographic information. They were required to play the entire audio file for a given voice and answer all questions about it before being able to move on to the next voice and moving backwards in the survey was not allowed. \ndedit{After participants heard each audio file, they rated the Likert items described above and answered the free-response question. Participants then repeated the process for the remaining voices. After all voices were rated and described, the study session ended.} 
The survey took approximately 5 minutes to complete and participants were compensated US \$1.25. \ndedit{We used the following inclusion criteria for participants recruited from MTurk: they were located in the United States, had an approval rate of 99\% or higher, and had performed at least 1,000 tasks previously.}

\subsection{Participants}
We recruited 65 participants through Amazon Mechanical Turk. All passed our inclusion criteria of fully answering qualitative questions, and thus no responses were excluded from analysis. We requested gender information through open-ended responses as recommended by HCI guidelines for collecting gender data. Open-ended responses reduce the negative experiences of participants who may not align with the check boxes provided \cite{scheurmangender2020}, and the coding process was tractable for this number of participants. The open responses were manually coded. Participants' ages ranged from 22 to 60 years, with a median age of 32. Participants self-identified as men (40), women (24), and non-binary (1). Their reported ethnicities were Asian (4), Black or African American (2), Biracial (1), Hispanic (2), Native American (1), and White (53). Six participants identified as part of the LGBTQ+ community. 

\subsection{Evaluation of Design Principles}\label{voice_results}

We used a combination of quantitative and qualitative methods to analyze the design principles described in Section~\ref{design_principles}.

\noindent\textbf{DP1: Pitch Modulates Gender Perception}\\
To evaluate gender perception of voice, we analyzed participants' responses of perceived masculinity and femininity. The responses showed an extremely high intrarater reliability when the scores of the masculinity scale were flipped/negated (Cronbach's $\alpha=.97$). While it is possible that non-human vocalizations can be perceived as not being gendered (e.g., chirps, beeps, or buzzes \cite{cha2018survey}), we observed that all voices in our study were perceived as gendered along an axis from masculinity to femininity. We posit that this is because all voices spoke in a human language, and were thus highly anthropomorphic. For this reason, we averaged the two items of perceived masculinity and perceived femininity into a single construct, arbitrarily choosing positive values to be feminine and negative values to be masculine. In the context of the voice study, we refer to this averaged construct as femininity.

Using this combined metric, we visualized the effect of fundamental frequency on the perceived gender of the voices, as shown in Figure \ref{fig:voice_frequencies}. Only one voice, Kendra, exhibited significantly different gender perceptions at different frequencies. Through a visual data analysis process \cite{szafir2021connecting}, we identified three regions (shown in Figure \ref{fig:voice_frequencies}) that were relatively consistent in femininity ratings and grouped the data into those ranges for analysis.

We performed a Welch ANOVA analysis to observe the effect of frequency group on perceived gender. We found that there was a significant main effect of frequency range on gender $F(2, 31.42)=37.64$, $p<.001, \eta_p^2=.46$. Using pairwise Games-Howell post-hoc tests revealed that the frequency range of 116Hz-128Hz ($M_{femininity}$=-.97) was perceived as having significantly lower perceived femininity than the frequency range 128Hz-143Hz ($M_{femininity}$=.75), $p=.003, \eta^2=.25$ which was, in turn, lower in perceived femininity than the frequency range 143Hz-164Hz ($M_{femininity}$=2.27) $p<.001, \eta^2=.65$. These results support \textit{DP1}, and narrowed our design space to one voice. We then verified that the Kendra voice fit the subsequent design principles for use in the integrative video study.

\noindent\textbf{DP2: Clarity is Maintained Across Genders}\\
To verify that the Kendra voice is both clear and realistic across the different perceived genders, we examined the realism and clarity items of our survey. For both items, we conducted Welch ANOVA analyses. With realism as a dependent variable, we observed no significant change between fundamental frequency ranges $F(2, 33.579) = 0.117$, $p=.890$. Similarly, with clarity as a dependent variable, we observed no main effect across fundamental frequency ranges $F(2,32.326)=2.41$, $p=.106$. From these two tests and our sample size, we find that there is no large disparity in clarity and realism between the range frequencies, making the Kendra voice a good choice for the integrative study.

\noindent\textbf{DP3: Identity Follows Function}\\
We also qualitatively evaluated the aspects of the Kendra voice to see if it would be useful for the context of our integrative study. We planned on selecting a voice that sounds professional, because the robot was being presented as a medical professional or a hotel receptionist. We found that participants described the voice as sounding educated and professional at different fundamental frequencies.  For example:

\textit{``This sounds like a robot who was programmed to sound like an average American educated male."} --P24, $f_0=131Hz$

\textit{``It sounds very formal and more business like but also still robotic."} --P16, $f_0=140Hz$

In addition to these responses, 25 of the 63 other participants similarly noted that the voice sounded ``robotic" when asked to describe the voice, despite the lack of indication that the voice would be used on a robot. This provides further evidence to support the fit of the selected Kendra voice for the context for which the voice would be used in the integrative study, satisfying our third design principle.

\begin{figure}[ht]
    \centering
    \includegraphics[width=.8\linewidth]{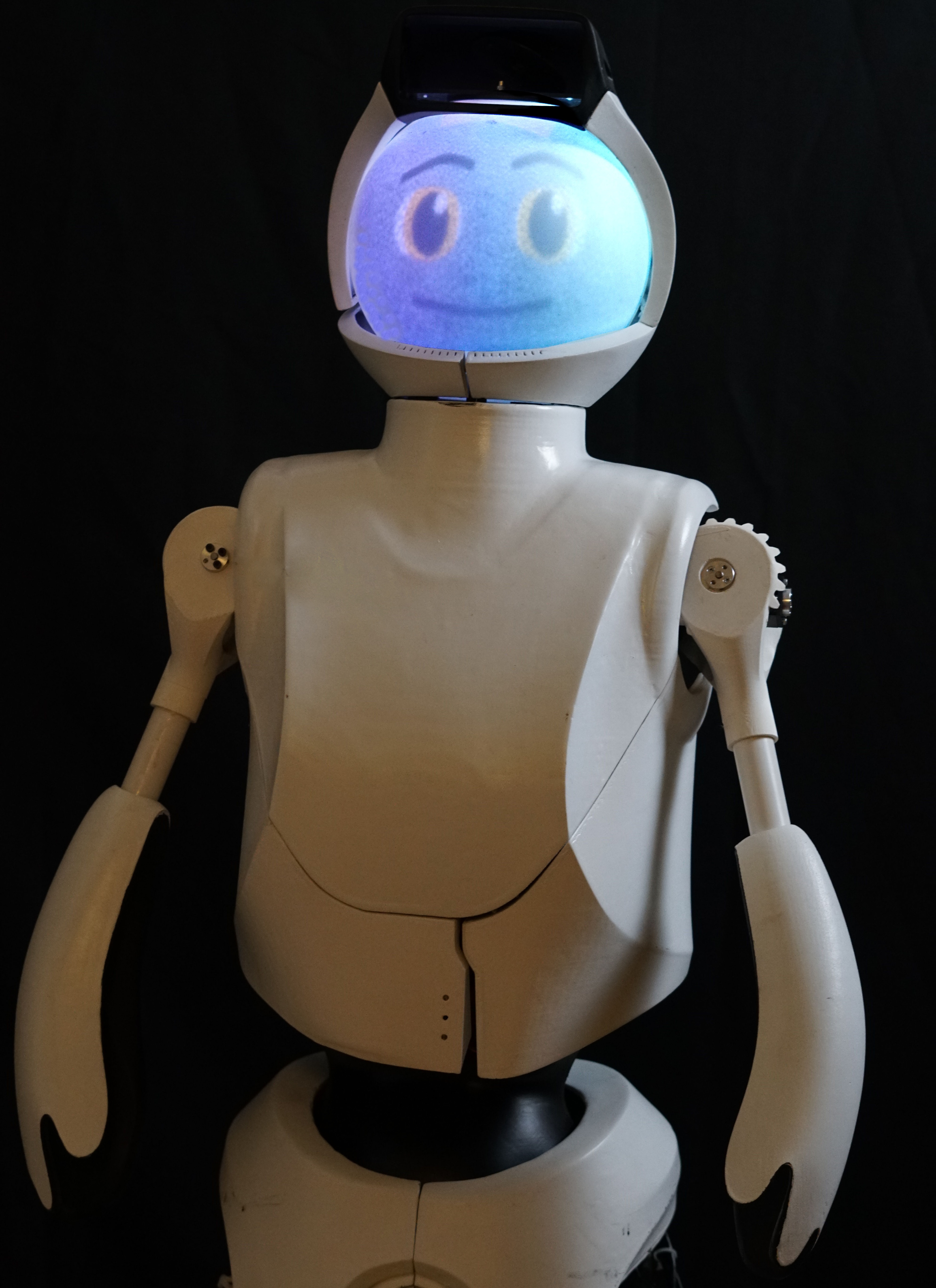}
    \caption{Quori (unclothed) \cite{specian2021quori}, the robot we selected to use for the clothing design study and integrative video study.}
    \label{fig:bare_quori}
\end{figure}

\section{Clothing Design Study}\label{clothing_study}
\ndedit{We were interested in understanding how appearance changes a person's perception of a robot's gender. Changing a robot's physical design is costly and requires significant expertise, making it inaccessible to users, especially those from marginalized populations. To make the design of appearance more accessible, we were inspired by the work of Friedman et al. \cite{friedman2021robots} that discussed how clothing can modify a robot's appearance to establish a robot's identity. Clothing is affordable, can be created with low-cost supplies, and users can learn to sew from tools and tutorials \cite{leake2023institches}.} To select the robot's appearance for the integrative study detailed in Section \ref{video_study}, we performed an evaluative design study of the perceived genders of a set of candidate clothing designs \ndedit{on their own}. \ndedit{To achieve this, we introduce a design methodology based on the four elements of fashion design \cite{elementsoffashiondesign}, applied to robots.} We selected the humanoid robot Quori \cite{specian2021quori} (shown in \autoref{fig:bare_quori}, because it was specifically designed to have a gender-neutral embodiment. We then designed clothing for the two task contexts \ndedit{planned for the integrative video study}: a medical professional and a hotel receptionist, \ndedit{because prior work has shown that the appearance of a robot should align with its intended task \cite{goetz2003matching}. The justification for selecting these tasks is provided in \autoref{video_study}.}

\subsection{Design Principles}\label{clothing_design_principles}
We set the following design principles (DPs) to craft the robot's appearance for the clothing design study:
\begin{itemize}
    \item \textbf{DP 1: Appearance Modulates Gender} 
    We aimed to design clothes that modulate the perceived gender of the robot. While clothing can be worn by any gender, the social construction of gender implies that the design of clothing is typically aligned with particular genders \cite{crane2012fashion}. For example, fashion designers tend to use curved lines in women's clothing and sharp lines in men's clothing. \cite{palumbo2015comparing}.
    
    \item \textbf{DP2: Quality and Cost are Consistent} Associations of different monetary values of clothes on the robot could lead to different social perceptions of that robot \cite{de2020leveraging}. The perceived quality and value of the robot's clothes should not make the robot appear more or less ``premium", nor should the clothes ``cheapen" the robot. 
    
    \item \textbf{DP3: Clothing Suggests Function} The clothes that a robot wears should resemble the clothes worn by people in similar occupations/contexts.  
    
\end{itemize}

\begin{figure*}[ht]
    \centering
    \includegraphics[width=\textwidth]{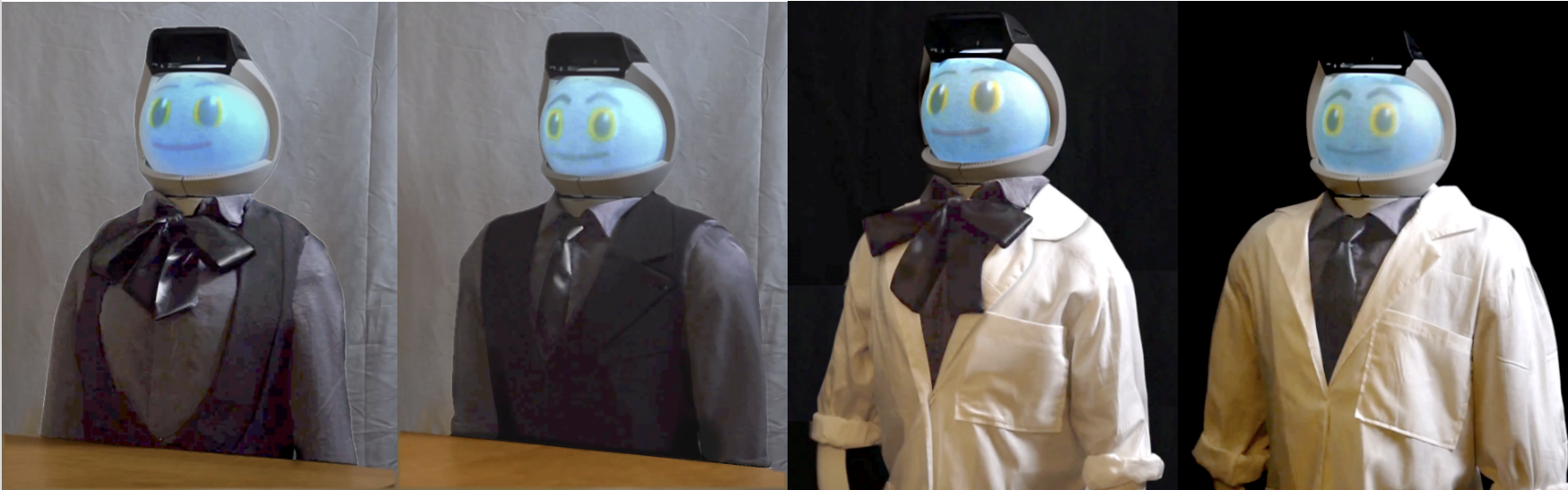}
    \caption{Appearance modifications of the Quori robot. The first two images represent the feminine and masculine versions of the robot clothing designed for the hotel receptionist task. The second two represent the feminine and masculine clothing designed for the medical professional task.}
    \label{fig:quori_appearance}
\end{figure*}

\begin{table*}
\centering
\caption{Clothing design results. We report the marginal means of femininity ($\mu_{feminine}$) and masculinity ($\mu_{masculine}$) across the different task and clothing types. We also report the number of participants that selected each clothing type as the most feminine ($N_{feminine}$) or the most masculine ($N_{masculine}$) when presented with all three clothing options for each task. Values following $\pm$ represent standard error.}
\label{tab:clothing_design_results}
\begin{tabular}{r|ccc|ccc}
\hline
Task & \multicolumn{3}{c}{Hotel Receptionist} & \multicolumn{3}{|c}{Medical Professional} \\
Clothing Type & Feminine & None & Masculine & Feminine & None & Masculine \\
\hline
$\mu_{feminine}$ & $.93 \pm .30$ & $.74 \pm .28$ & $-.78 \pm .23$ & $.50 \pm .26$ & $.86 \pm .34$ & $-.81 \pm .23$ \\
$N_{feminine}$ & 40 & 51 & 2 & 50 & 41 & 2 \\
$\mu_{masculine}$ & $-.89 \pm .30$& $-1.33 \pm .24$ & $.70 \pm .17$ & $-1.04 \pm .26$ & $-1.79 \pm .27$ & $.64 \pm .19$ \\
$N_{masculine}$ & 4 & 3 & 86 & 3 & 4 & 86 \\ \hline
\end{tabular}
\end{table*}

\subsection{Design Choices}
To adhere to our design principles, we made the following design choices in constructing the robot's clothes.  \ndedit{We aimed to evaluate perceptions of robot genders in the United States, and thus our design process reflects these Western and American patterns and gendered interpretations of clothing design.} We designed a dress shirt, a lab coat, and a vest, each in a current American masculine and a current American feminine style, as shown in Figure~\ref{fig:quori_appearance}. By putting these clothes on the robot, we investigate how these clothing design choices may be \textit{performative}, and establish the robot's identity. Using our design process, designers with multiple intersecting identities can explore how their \textit{unique standpoint} can inform choices in the design space of robot appearance and construct knowledge.

\textbf{Silhouette:}
To align with \textit{DP1}, we modulated the silhouette of the clothing to suggest different genders by using quilt batting, inspired by drag queens \ndedit{on RuPaul's Drag Race} who use padding to alter their bodies to appear more feminine \cite{darnell2017werk}. Quilt batting offers a light-weight and cost-efficient method to modulate the underlying embodiment of the robot since careful placement can have the effect of augmenting the robot's embodiment without impacting function. Batting has the additional property of being highly compliant, contributing to safety in physical interaction scenarios.

For the masculine silhouette, we added several layers of batting to the shoulders and arms to give the robot a larger look representative of findings of perceived robot masculinity in prior work \cite{trovato2018she}. We added darts to the feminine style to reduce the width of the shoulders to give the garment a less ``boxy" shape, reflective of typical Western women's clothes \cite{mccunn2016sewing}.

\textbf{Lines:}
We modulated the perceived gender of the robot by changing the style lines of the clothes, with feminine clothes containing more curved lines and masculine clothes containing more straight and angled lines. We incorporated two main manipulations: style of accessory and shaping of the garment's lapels and collar.

For the accessory, we used a tie for the masculine robot, which established straight and angular lines in the garment. For the feminine robot, we used a lavalli\`ere as the accessory. Because the lavaill\`ere  droops due to gravity, the lines it establishes are naturally curved. These accessories were chosen because they are typically used in formal Western contexts and are not associated with specific eras in Western fashion.

Furthermore, we altered the collar of the dress shirt and lapels of both the lab coats and vests. For the masculine appearance, the collars and lapels were sewn into a sharp point. For the feminine appearance, the collar and lapels were instead sewn with a curved line. For the vest, we dropped the opening of the feminine version to create a more curved style line.

\textbf{Color and Texture:}

To follow \textit{DP2}, we used the same fabrics for both garments of each type, since fabric contributes to a garment's perceived price and quality \cite{swinker2006understanding}. For the dress shirt, we chose a grey taffeta fabric to reflect the natural sheen of the unclothed robot surfaces and a gender-neutral color and texture to be worn underneath other clothes. We chose black silk for the neck accessories, because is traditionally used for Western neck ties and bows. We used a white cotton blend fabric for the lab coats, consistent common Western medical lab coats. The vest was made from a black stretch denim fabric, giving a stiff appearance and structure. To control for quality, a single person (the first author) constructed all the garments using the same methods and materials.

\textbf{Face Design:}
We used the robot's face as a natural indicator of gender. Drawing on studies in human facial perception, we created a feminine face by using large eyes, thin eyebrows, and red-tinted lips \cite{mogilski2018relative}. In contrast, the masculine face had thicker eyebrows and smaller eyes with grey-tinted lips. The neutral unclothed robot had a face that was perceptually exactly in the middle between the masculine and feminine face in terms of size of features and lip color. The face was implemented with the PyLips Python package \cite{dennler2024pylips}. 

\subsection{Study Description}

\ndedit{To evaluate the perceived gender of our clothing design, we performed an online user study via Amazon Mechanical Turk that followed a mixed design. Each participant was presented with static images of the robot to rate, similar to the images in \autoref{fig:quori_appearance}. As in the voice study, these images were presented without context, i.e., the robot did not have a name or voice, and the intended context was not described to the participants. The within-subjects factor of this study was the task that the robot was designed for: the hotel receptionist and medical professional tasks, and the between subjects factor were the gendered qualities of the clothing: masculine, feminine, or androgynous (the unclothed robot, as in \autoref{fig:bare_quori}).} These conditions were randomized and counter-balanced. Participants in the clothing study filled out an on-line questionnaire also administered through Amazon Mechanical Turk.

For each stimulus, participants \ndedit{responded to the following seven-point Likert items ranging from ``Strongly Disagree" to ``Strongly Agree":}
\begin{enumerate}
    \item This robot seems masculine (masculine)
    \item This robot seems feminine (feminine)
    \item This robot seems expensive (cost)
    \item This robot seems high-quality (quality)
\end{enumerate}

\ndedit{After responding to these questions, participants were asked to respond to the open-response prompt ``What are one to two tasks you think that the robot would be capable of doing?". After these responses were collected for the first task, }participants were presented with all three clothing options for the first task and asked to choose the most masculine option and the most feminine option. We also collected open-ended responses to the prompt: ``Briefly describe your reasoning for selecting that robot" after each selection of the most masculine and feminine option for that task. \ndedit{After that, the participant followed the same process for the other task and then ended the study session.} The study was deployed in September 2021. The survey took approximately 5 minutes to complete and participants were compensated with US \$1.25. \ndedit{We used the following inclusion criteria for participants recruited from Amazon Mechanical Turk: they were located in the United States, had an approval rate of 99\% or higher, and had performed at least 1,000 tasks previously.}

\subsection{Participants}

We recruited 100 participants for our clothing study, approved under USC IRB \#UP-18-00510, through Amazon Mechanical Turk. We eliminated the data of participants who failed our attention check, which resulted in 93 valid data points. A chi-square test reveals that this did not affect the underlying distribution of conditions, $\chi^2(2,n=93)=2.77$, $p=.250$. \ndedit{We used the same coding process for gender and ethnicity as in the voice design study.} The final set of participants consisted of 47 men, 45 women, and 1 non-binary person. The ethnicities of the participants were: Asian (4), Black or African American (5), Hispanic (4), Native American (2), 4 Multiracial (4), and White (71). Eight participants reported that they identified as part of the LGBTQ+ community.

\subsection{Study Results}\label{clothing_results}
We used a mixed-methods approach to evaluate the design principles described in Section~\ref{clothing_design_principles}.

\noindent\textbf{DP1: Appearance Modulates Gender}\\
To evaluate the perceived gender of the robots, we investigated the effects of visual stimuli on the perceived masculinity and femininity. The correlation between masculinity and femininity was much lower in this study than in the voice study (Cronbach's $\alpha=.73$), thus we analyzed the two items independently. We also considered the two tasks (medical professional and receptionist) independently. We found a main effect of clothing condition on perceived femininity, $F(2, 51.67)=14.76$, $p<.001$ for the medical professional task as well as the receptionist task $F(2, 58.09)=13.62$, $p<.001$, using Welch's ANOVA due to heteroscedasticity of variances. The effect of the clothing condition on masculinity was also significant for both the medical professional task $F(2, 56.15)=26.97$, $p<.001$ and the receptionist task $F(2, 55.06)=28.12$, $p<.001$. \ndedit{The marginal means for femininity and masculinity ratings are shown in \autoref{tab:clothing_design_results}. Post-hoc analysis using Tukey's test revealed that the feminine medical professional design was perceived as both more feminine, $p=.003$, and less masculine, $p=.001$, than the masculine medical professional design. Similarly, the feminine receptionist design was perceived as both more feminine, $p=.001$, and less masculine, $p=.001$, than the masculine receptionist design.}

Interestingly, there were no significant differences in femininity or masculinity of the unclothed ``neutral" robot compared to the femininely-dressed robots in either \ndedit{the hotel receptionist task, $p=.886$, or the medical professional task, $p=.603$}. This was additionally reflected in the choice condition, where participants were equally likely to choose the unclothed robot or the femininely dressed robot as being the most feminine. When asked for their reasoning, many participants cited both the silhouette and style lines of the robot. For example, participants indicated the the unclothed robot was the most feminine because it was ``slimmer" than the other robots, while other participants described the feminine clothes as ``...the least bulky". With respect to lines, participants noted that the unclothed robot ``has curves compared to the other two" whereas other participants mentioned that the feminine clothes were more feminine because the robot had ``...a rounded collar and a flouncy bow". This shows that \textit{DP1} was correctly implemented \ndedit{with respect to the clothing designs, but calls into question whether the goal of an androgynous form was achieved in the Quori design}.

\noindent\textbf{DP2: Quality and Cost are Consistent}\\
We performed Welch ANOVAs independently for the two different tasks with the clothing conditions as a between-subjects variable. For the medical professional task, we observed \textit{no significant effect} of clothing on perceived quality, F(2,59.14)=.68, $p=.370$, or perceived cost, F(2,58.28)=1.09, $p=.344$. Similarly, for the receptionist task, we found \textit{no significant effect} of clothing on perceived quality, F(2, 56.47)=.46, $p=0.634$ or perceived cost, F(2,54.56)=1.01, $p=0.371$. Given the size of this study, it is unlikely that there is a large difference in the perceived quality and cost, supporting that design principle \textit{DP2} was successfully implemented.	

\noindent\textbf{DP3: Clothing Suggests Function}\\
We qualitatively evaluated that the clothes matched the function of the task by \ndedit{performing a deductive thematic analysis based on the generated categories of robot jobs from previous works \cite{dennler2022using,kalegina2018characterizing}.  Specifically, we were interested in evaluating the ``medical" and ``customer service" categories, to reflect the two tasks we planned for the integrative study}. For the medical professional task, we counted the responses that mentioned medical domains. Four of the 28 responses for the feminine clothing condition and 8 of the 36 responses for the masculine clothing condition mentioned that the robot could work in the medical field. For the unclothed condition, however, only 1 participant out of 29 believed that the robot was capable of performing medical tasks. The hotel receptionist task was characterized by working in a customer service domain; 13 of the 27 responses for the masculine clothing and 16 of the 27 responses for the feminine clothing described the robot as doing customer service tasks. Only 8 of the 39 responses for the unclothed condition described the same context. The higher proportion of responses specific to the actual task indicates that \textit{DP3} was achieved.

\section{Integrative Video Study}\label{video_study}

Building on the findings from the two studies described so far, we next performed an online video-based integrative study to explore how task, voice, and appearance interact to form gender perceptions and other social perceptions of a robot. \ndedit{To evaluate the impact of the task, we selected tasks of two varying social roles because previous research and several ethical design frameworks highlight the importance of power dynamics between the user and the robot \cite{winkle2023feminist,zhu2024robots}. We selected the hotel receptionist task as a task where the robot has a lower social role than the user, and we selected the medical professional task as a task where the robot has a higher social role than the user. These tasks have employed an approximately equal numbers of men and women \cite{campos2011patterns,pelley2020specialty}, however, people also hold gendered stereotypes about these tasks, such as that receptionists are more feminine and medical professionals are more masculine \cite{bryant2020should}.} Participants viewed a video of the robot as a medical professional that \textit{provided} instructions to a ``patient" (an actor), and as a hotel receptionist that \textit{received} instructions from a ``patron" (an actor).

\begin{figure}
    \centering
    
    \begin{subfigure}[b]{\linewidth}
    \includegraphics[width=\linewidth]{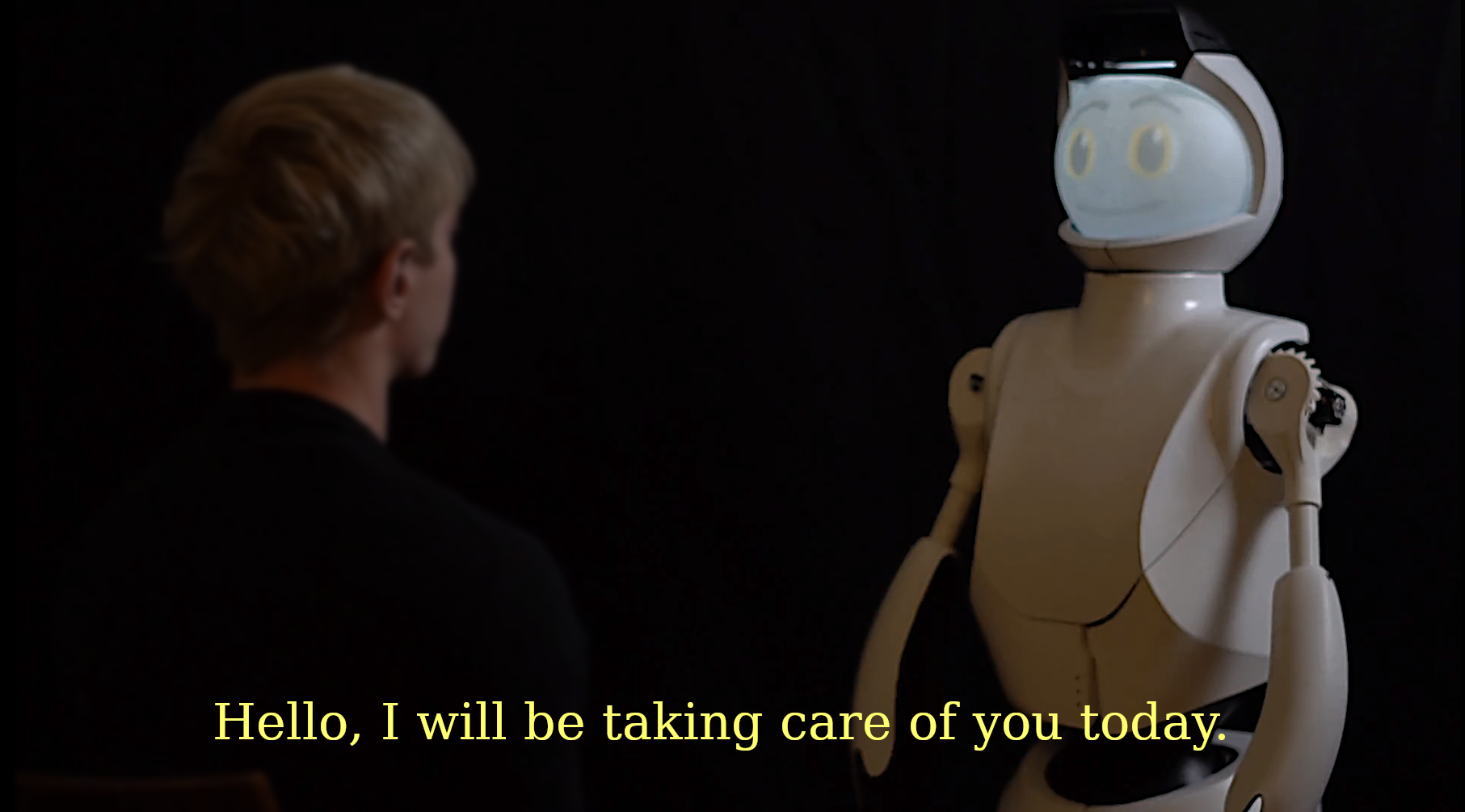}
    \vspace{0cm}\hfill
    \includegraphics[width=\linewidth]{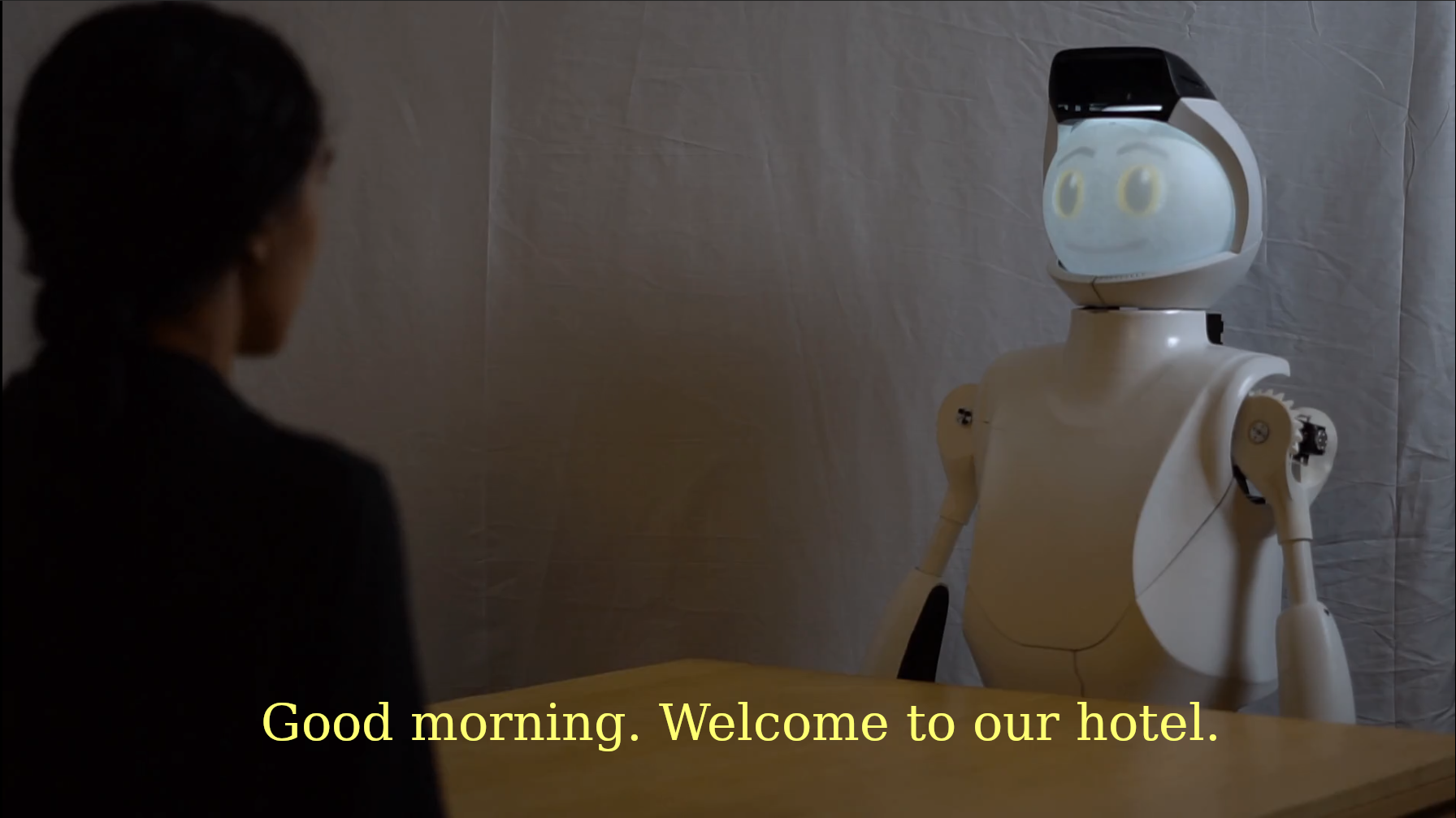}
    \vspace{0.4cm}
    
    \end{subfigure}\hfill

    \caption{Sample frames from the two tasks we selected: medical professional (top) and hotel receptionist (bottom).}
    \label{fig:frames}
\end{figure}
\subsection{Hypotheses}
From prior research in human and robot gender perception, as well as our two studies described above, we developed the following hypotheses regarding  social perceptions of the robot as related to task and gender:

\begin{itemize}
    \item \textbf{H1:} \ndedit{\textit{Participants will ascribe different social attributes to the robot depending on the task the robot performed in the video (medical professional vs. receptionist).}}
    \item \textbf{H2:} \ndedit{\textit{Participants will ascribe different social attributes to the robot depending on the the perceived gender of the robot that is performing a specific task.}}
    \item \textbf{H3:} \ndedit{\textit{Participants will ascribe different social attributes to the robot when the robot's gendered cues are aligned compared to when the robot's gendered cues are different.}}
\end{itemize}

Additionally, we developed two more hypotheses regarding differences in the perceived gender of the robot, one regarding the robot's perceived gender as it relates to its appearance and voice, and the other regarding the effect of task-aligned clothing on the robot's perceived gender in the absence of strong gender cues ~\cite{friedman2021robots}:
\begin{itemize}
     \item \textbf{H4:} \ndedit{\textit{Participants will assign different genders to the robot depending on the voice \ndedit{(H4a)}, appearance \ndedit{(H4b)}, and the interaction of voice and appearance \ndedit{(H4c)}.}}
     \item \textbf{H5:} \ndedit{\textit{Participants will assign different genders to the robot depending on the appearance \ndedit{(H5a)}, task \ndedit{(H5b)}, and their interaction \ndedit{(H5c)} when the robot has an ambiguously-gendered voice.}}
\end{itemize}

\begin{figure*}[ht]
\centering
\begin{subfigure}[b]{0.2\textwidth}
    \includegraphics[width=\linewidth]{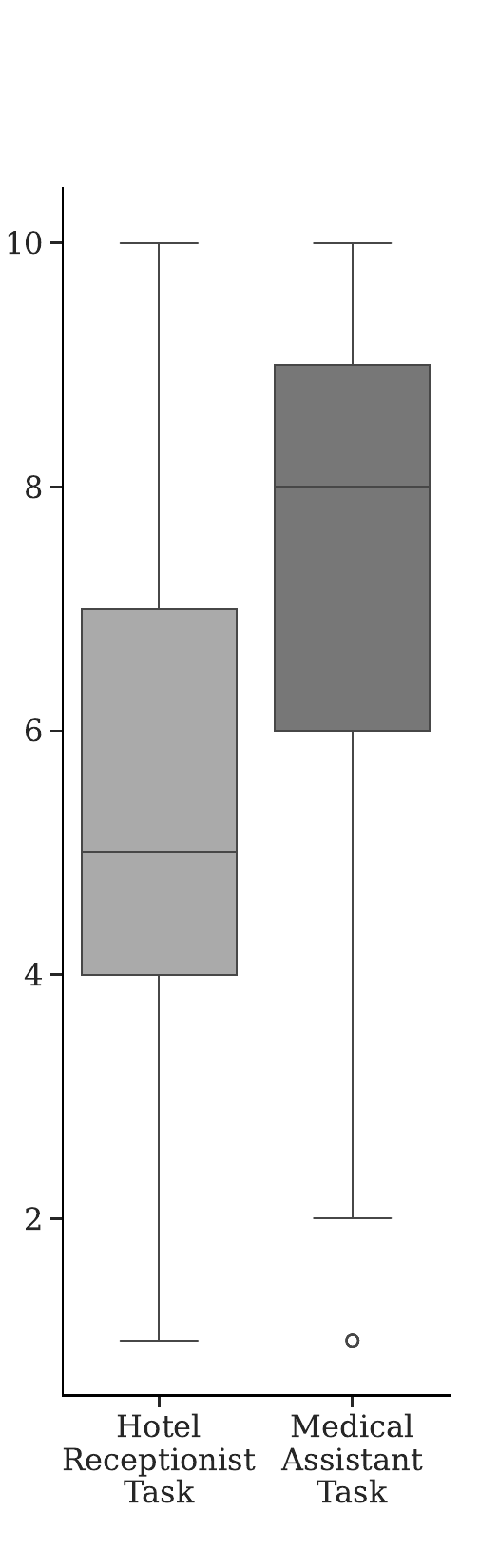}
    \caption{Perceived social role of the two tasks.}\label{fig:role}
\end{subfigure}\hfill
\begin{subfigure}[b]{0.38\textwidth}
    \includegraphics[width=\linewidth]{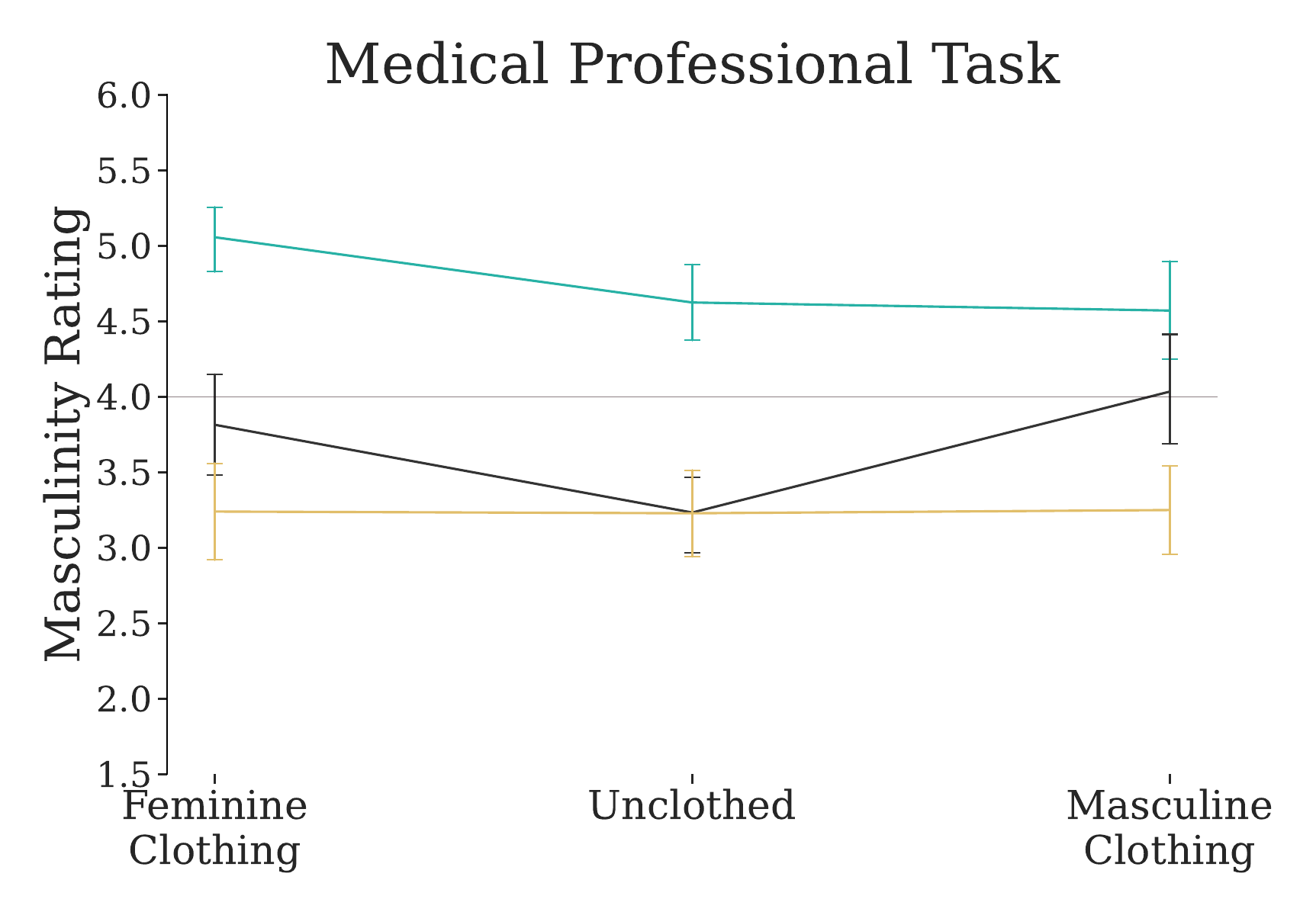}
    
    \includegraphics[width=\linewidth]{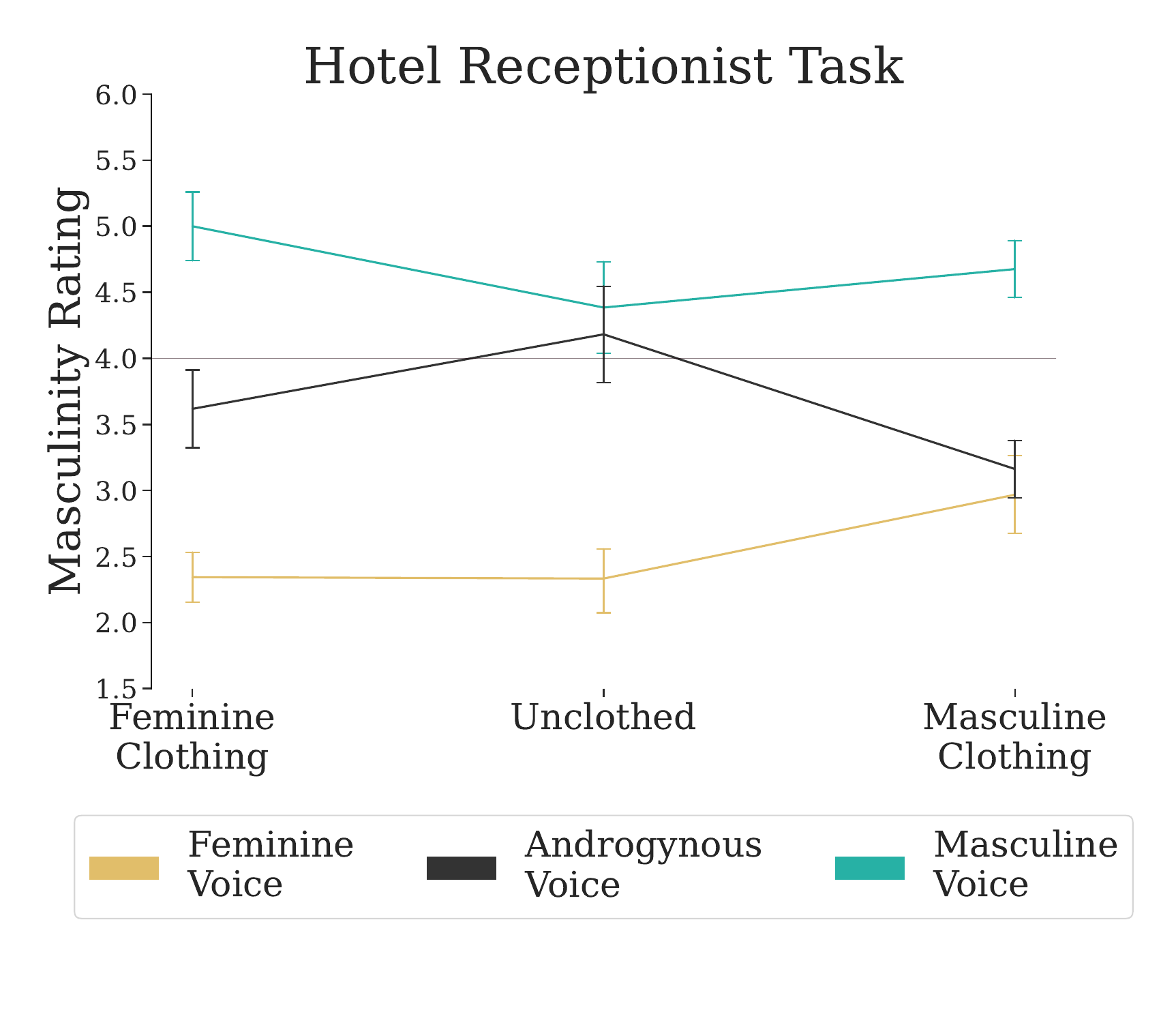}
    \caption{Masculinity ratings}\label{fig:masculinity}
\end{subfigure}\hfill
\begin{subfigure}[b]{0.38\textwidth}
    \includegraphics[width=\linewidth]{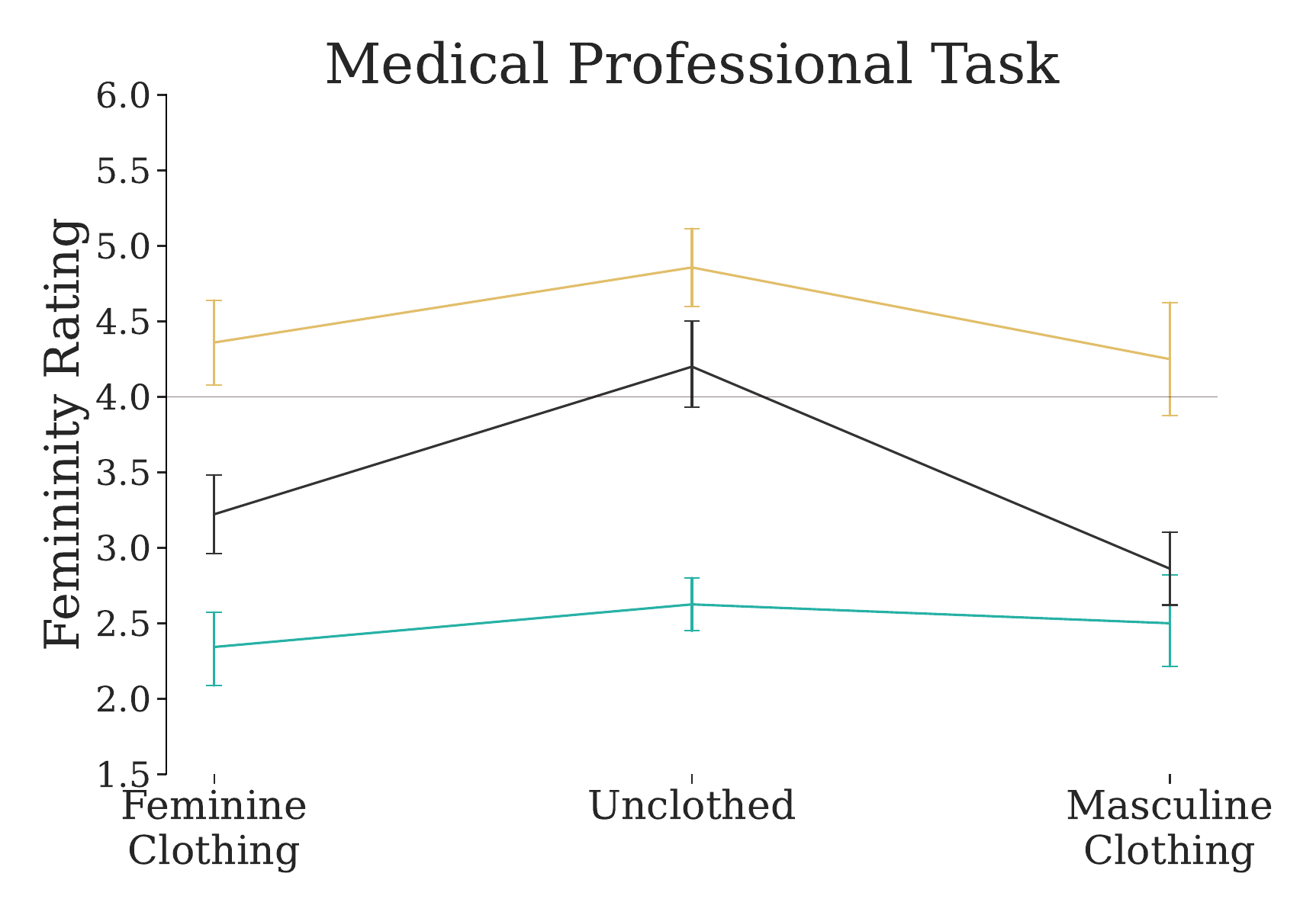}
    
    \includegraphics[width=\linewidth]{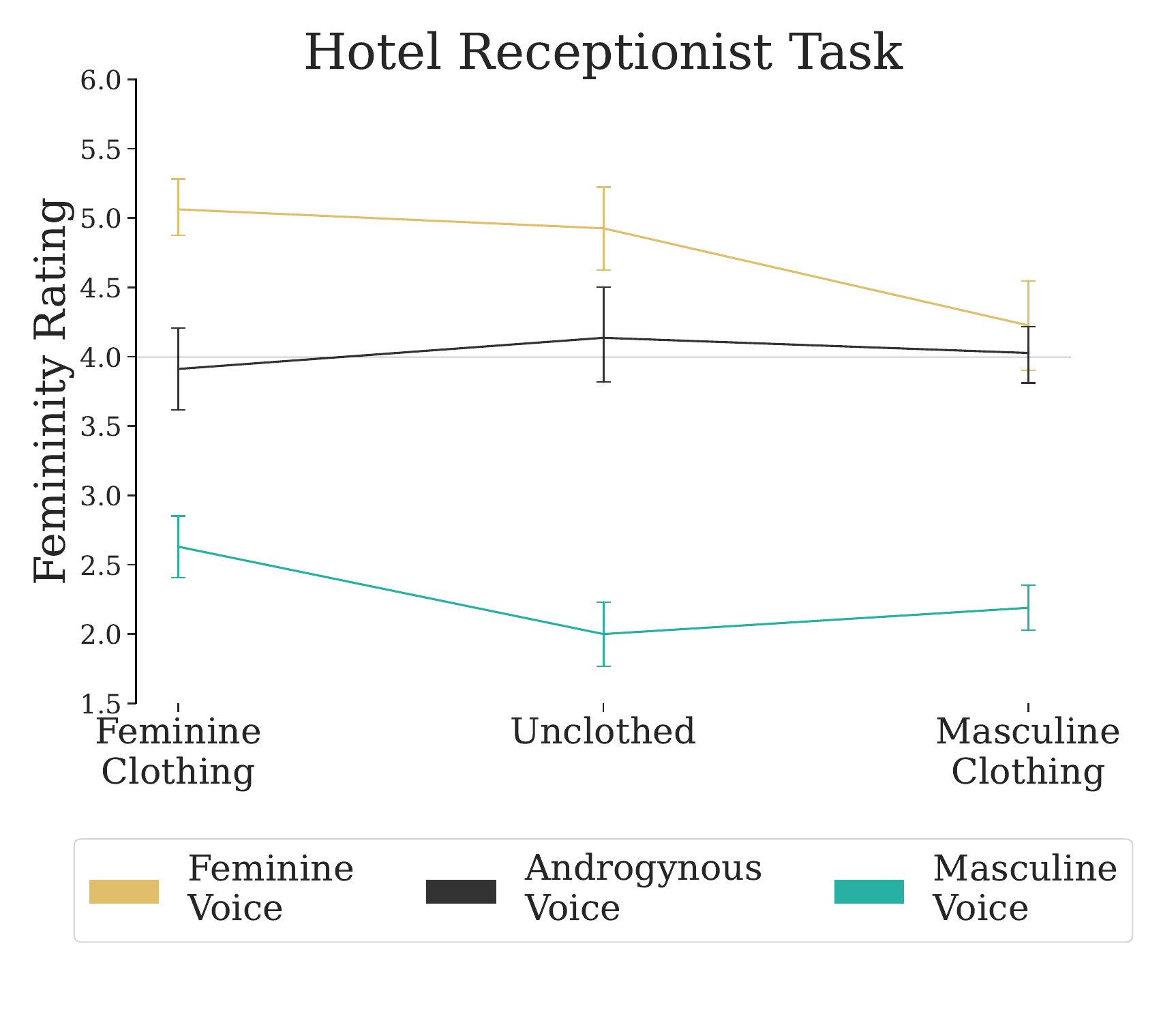}
    \caption{Femininity ratings}\label{fig:femininity}
\end{subfigure}
    \caption{Summary of stimuli and results from the integration study.  We found that the manipulation of the social role was significantly different between the two conditions ($p<.001$), with the expected social role of the receptionist task being lower than the social role of the medical professional task (a). Participants saw videos of the robot performing a task with a human. We also found that voice and appearance affected the perception of masculinity (b) and femininity (c) in both conditions.}
    \label{fig:quori_videos}
\end{figure*}

\subsection{Manipulated Variables}
We manipulated three robot variables in this study: task, voice, and appearance. Task was manipulated within-subjects following a random and counter-balanced design, and voice and appearance were manipulated between subjects.

\begin{itemize}
    \item \textbf{Task}: There were two levels of task: hotel receptionist and medical professional, based on the considerations described in \autoref{video_study}.
    \item \textbf{Voice}: There were three levels of voice that we created by modifying the default Kendra voice from Amazon Polly, in accordance with the results detailed in Section \ref{voice_results}: feminine (up one semitone), androgynous (down one semitone), and masculine (down three semitones).
    \item \textbf{Appearance}: There were three levels of appearance following the results detailed in Section \ref{clothing_results}: unclothed, feminine clothing, and masculine clothing. 
\end{itemize}

\subsection{Study Description}
Participants filled out an online questionnaire deployed on Amazon Mechanical Turk. They began by reporting their demographic information and then filled out the Negative Attitude toward Robots Scale (NARS) \cite{nomura2006measurement}. The participants were then shown a 90-second video (Figure \ref{fig:frames}) of a robot performing one of the two tasks with a randomly assigned voice and appearance from the validated options described above. \ndedit{The choice of voice and appearance were randomized and counter-balanced in a 3x3 matrix. In the medical professional task, the video showed the robot performing routine medical tests on the actor: asking the actor to show their arm to collect pulse information and make different faces to test cranial nerve function. In the hotel receptionist task, the robot was directed by the actor to make changes to a reservation. Each videos was approximately 90 seconds long. After the participants viewed a video, they} rated the perceived social role of the robot in the video on a Likert scale from 1 to 10. The participants then rated the robot on the Robotic Social Attributes Scale (RoSAS) \cite{carpinella2017robotic}. We then asked participants to use a short open-response dialogue box to describe what the robot was doing, what they would name the robot, and what other tasks they expected the robot to do. The participants were then shown a 90-second video of the robot performing the other task and answered the same questions. The study was deployed in September 2021. The survey took approximately 10 minutes to complete and participants were compensated with US \$2.50. \ndedit{We used the following inclusion criteria for participants recruited from Amazon Mechanical Turk: they were located in the United States, had an approval rate of 99\% or higher, and had performed at least 1,000 tasks previously.}

\subsection{Participants}
We recruited 360 participants via Amazon Mechanical Turk. After removing responses that failed the attention check, we were left with 273 valid responses. A chi-square test revealed that this exclusion did not significantly alter our random assignment of voice and appearance, $\chi^2(4, n=273)=2.01$, $p=.733$. The final study population consisted of 149 men, 118 women, 3 non-binary people. The ethnicities of the participants were: 20 Asian (20), Black or African American (16), Hispanic (8), Multiracial (7), and White (217). In addition, 38 participants reported that they identified as part of the LGBTQ+ community. 

We observed acceptable internal consistencies between all constructs measured by the NARS ($\alpha_{\text{future influence}} = .87$, $\alpha_{\text{relational attitudes}} = .80$, $\alpha_{\text{actual interaction}} = .78$) and RoSAS ($\alpha_{\text{warmth}} = .79$, $\alpha_{\text{competence}} = .82$, $\alpha_{\text{discomfort}} = .78$) scales.

\subsection{Manipulation Check}
To evaluate that our selection of tasks was effective, we performed a manipulation check of the reported social roles of the participants for each task. We applied a non-parametric Wilcoxon Signed-Rank test and found that the medical professional task had a significantly higher social role than the hotel receptionist task, W=2282.5, $p<.001$, as shown in Figure \ref{fig:role}. This indicates that our manipulation of the tasks' social role was successful.

\subsection{Main Study Results}

We evaluated our hypotheses related to both social perception of the robot and the perceived gender of the robot.

\subsubsection{Social Perception}
\noindent\textbf{H1: Tasks Affect Social Perception}: Our first hypothesis postulated that different tasks affect the social perception of the robot. We evaluated this hypothesis using a non-parametric Wilcoxon signed-rank test between task conditions. We found that the medical professional was rated as significantly \textit{less warm}, W=7250, $p<.001$, \textit{less competent}, W=3701, $p<.001$, and \textit{more discomforting}, W=5949.5, $p<.001$ than the hotel receptionist robot. Essentially, the robot with the higher social role was rated as less socially favorable. Therefore, \textbf{H1} is supported.\\

\noindent\textbf{H2: Gender Perception Affects Social Perception}: The \ndedit{second} hypothesis postulated differences in the RoSAS constructs of Warmth, Competence, and Discomfort. Naturally, these social evaluations are also dependent on participants' overall preconceptions of robots, as measured by the NARS constructs of Future Influence, Relational Attitudes, and Actual Interaction. Thus we analyzed the effect of robot gender on Warmth, Competence, and Discomfort using a two-way ANCOVA for each task with voice and appearance as between-subject variables and the subscales of Future Influence, Relational Attitudes, and Actual Interaction as covariates.

In the medical professional task, there were no main or interaction effects of voice and appearance on the RoSAS constructs of Warmth, Competence, or Discomfort. The NARS subscale Future Influence was significant for all three RoSAS constructs: Warmth ($F(1,261)=26.55$, $p<.001$), Competence ($F(1,261)=11.32$, $p=.001$), and Discomfort (F(1,261)=16.11, $p<.001$). Additionally, the Actual Interaction subscale of NARS was a significant predictor of all three RoSAS constructs: Warmth ($F(1,261)=88.92$, $p<.001$), Competence ($F(1,261)=10.61$, $p=.001$), and Discomfort ($F(1,261)=7.25$, $p=.008$). The Relational Attitudes were not significant for any RoSAS constructs.

Within the receptionist task, we found no significant main or interaction effects of voice and appearance on Warmth, Competence, or Discomfort. We did find, however, that the covariates were significant predictors of the RoSAS constructs. Future Influence was a significant predictor of Warmth ($F(1,261)=16.88$, $p<.001$), Competence ($F(1,261)=28.97$, $p<.001$), and Discomfort (F(1,261)=48.06, $p<.001$). Additionally, the Actual Interaction subscale of NARS was a significant predictor of Warmth ($F(1,261)=72.663$, $p<.001$) and Competence ($F(1,261)=6.61$, $p=.011$), but not Discomfort. As in the medical professional task, Relational Attitudes was not a significant predictor for any RoSAS construct. Therefore, \textbf{H2} was not supported.\\

\noindent\textbf{H3: Alignment of Cues Affects Social Perception}: Based on previous work that found that unaligned aesthetic cues lead to ambiguity \cite{paetzel2016effects}, we examined how the alignment of gendered cues may affect user social perceptions. We considered masculine voice with masculine appearance and feminine voice with feminine appearance as aligned cues, and masculine voice with feminine appearance and feminine voice with masculine appearance as unaligned, as in previous work \cite{mitchell2011mismatch}, noting that while this reflects normative views of gender, such normative views are instrumental in understanding how stereotypes may affect the construction of gender in robots. We did not consider androgynous cues as they are aligned with both masculine and feminine cues. We analyzed the effect of robot gender on the dependent variables Warmth, Competence, and Discomfort using a one-way ANCOVA for each task. The between-subjects variable was the alignment of gendered cues and the covariate was the NARS subscales. 

In the medical professional task, there were no main or interaction effects of alignment on the RoSAS constructs of Warmth, Competence, or Discomfort. The NARS subscale Future Influence was significant for all three RoSAS constructs: Warmth ($F(1,107)=8.65$, $p=.004$), Competence ($F(1,107)=7.60$, $p=.007$), and Discomfort ($F(1,107)=11.23$, $p=.001$). Additionally, the Actual Interaction subscale of NARS was significant for only Warmth ($F(1,261)=88.92$, $p<.001$). The Relational Attitudes were not significant for any RoSAS constructs.

In the receptionist task, there were again no main or interaction effects of gender cue alignment on the RoSAS constructs of Warmth, Competence, or Discomfort. The NARS subscale Future Influence was significant for all three RoSAS constructs: Warmth ($F(1,122)=14.11$, $p<.001$), Competence ($F(1,122)=8.70$, $p=.004$), and Discomfort ($F(1,122)=11.25$, $p=.001$). Additionally, the Actual Interaction subscale of NARS for only Warmth ($F(1,122)=32.13$, $p<.001$). The Relational Attitudes were not significant for any RoSAS constructs. Therefore, \textbf{H3} was not supported.

\begin{figure}[ht]
\centering
\begin{subfigure}[b]{0.4\textwidth}
    \includegraphics[width=\linewidth]{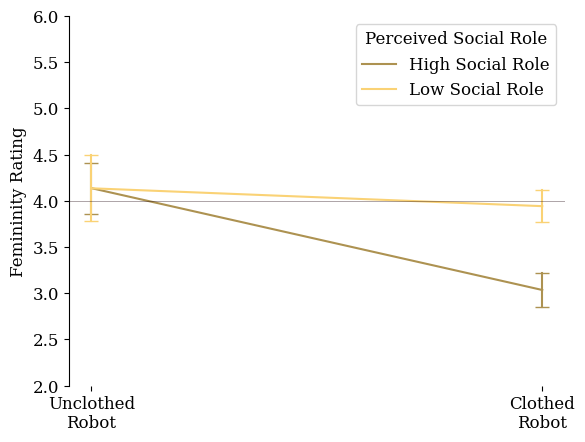}
    \caption{Femininity ratings by robot clothing for the androgynous voice robot.}\label{fig:femme_clothes}
\end{subfigure}
\begin{subfigure}[b]{0.4\textwidth}
    \includegraphics[width=\linewidth]{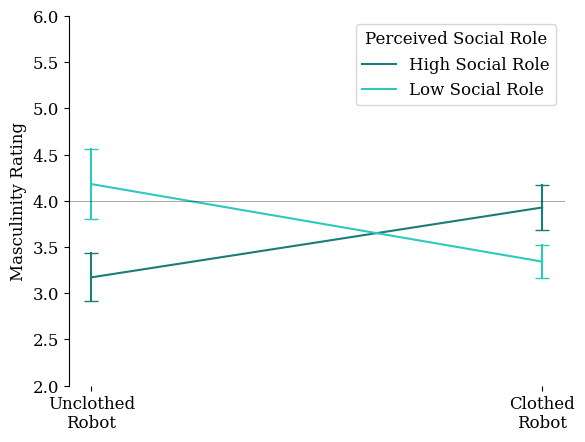}
    \caption{Masculinity ratings by robot clothing for the androgynous voice robot.}\label{fig:masc_clothes}
\end{subfigure}
    \caption{Summary of the effect of clothing the robot in the two tasks, for robots with androgynous voices. We found that clothing had a significant effect on gendering the robot, depending on the task. In tasks with higher social roles, clothing makes the robot more masculine and less feminine, and in the lower social role task, clothing makes the robot less masculine.}
    \label{fig:video_results}
\end{figure}

\subsubsection{Gendering Robots}
\noindent\textbf{H4: Voice and Appearance Affect Perceived Gender}: Our fourth hypothesis posited that voice and appearance affect the robot's perceived gender. We analyzed this independently for each task with a two-way ANOVA. We performed this separately for both ratings of femininity and masculinity. However, intrarater reliability for these two items was low (Cronbach's $\alpha=.53$), indicating that these scales are not likely to measure the same construct.

For the medical professional task, we observed a significant main effect of voice on femininity $F(2,264)=43.71$, $p<.001$ with a large effect size ($\eta_p^2=.249$). We also observed a main effect of appearance $F(2,264)=5.61$, $p=.004$, however the interaction was not significant. For masculinity there was a main effect of voice $F(2,264)=21.60$, $p<.001$ with a modest effect size ($\eta_p^2=.141$). The main effect of appearance and the interaction of appearance and voice were not significant for ratings of masculinity in the medical professional condition.

In the hotel receptionist task, we also observed a significant main effect of voice on femininity, $F(2,264)=71.85$, $p<.001$, with a very large effect size ($\eta_p^2=.352$), however, the main effect of appearance and the interaction effect of voice and appearance were not significant. For the perception of masculinity, voice had a significant effect, $F(2, 264)=45.34$, $p<.001$, again with a large effect size ($\eta_p^2=.256$). The main effect of appearance was not significant, but the interaction of voice and appearance was significant, $F(4, 264)=3.02$, $p=.018$. \ndedit{These results support \textbf{H4a}, do not support \textbf{H4b}, and partially support \textbf{H4c}.}


\noindent\textbf{H5: Identity Established through Clothing Affects Perceived Gender for Robots with Androgynous Voices}: We examined how reinforcing the task context with clothing as opposed to a generic robot body affects gender perceptions of the robots. Due to the larger effect of voice in the process of gendering robots that we saw in our previous studies, we investigated how clothing affects the perceived gender of robots with androgynous voices. To accomplish this goal, we performed a two-way ANOVA with the between-subjects factor being the role of the task, and whether or not the robot was wearing clothes, with masculinity and femininity ratings as the dependent variable.

We found that clothing did affect the perceived gender of the robot. Both the social role of the task ($F(1,173)=8.49$, $p=.004$) and whether or not the robot was wearing clothes ($F(1,173)=7.67$, $p=.006$) affected ratings of femininity with a small to medium effect ($\eta_p^2=.047$ for role, and $\eta_p^2=.042$ for clothing). The interaction of role and clothing was significant for ratings of masculinity with a small to medium effect size ($F(1,173)=8.48$, $p=.004$, $\eta_p^2=.047$). \ndedit{Therefore, \textbf{H5a}, \textbf{H5b}, and \textbf{H5c} are supported.}

\section{Discussion} \label{discussion}

In this work, we investigated the construction of robot gender identity through three user studies, two of which analyzed the design of the individual modalities to convey gender, and the third, integrative study examined the interaction between the two modalities, providing novel insights for the field \ndedit{of robotics}.


\subsection{\ndedit{Voice Design Implications}}

\noindent\textbf{\ndedit{Factors beyond pitch influence gender perception.}} Our synthetic voice design study found that pitch was a significant indicator of the perceived gender of some voices through evaluating \textbf{DP1} \ndedit{in the voice study}, consistent with some other studies~\cite{puts2006dominance,pernet2012role}. However, we also found that pitch was not the only indicator of gender, when identical fundamental frequencies between different voices were rated as having different perceived genders. \ndedit{For example, at an $f_0$ of 125Hz, Kendra, Joey, and Matthew were perceived as masculine; Joanna and Kimberly were perceived as ambiguously gendered; and Sally was perceived as feminine}. This suggests important questions about the other qualities of voices that can form perceptions of gender, and how those attributes can be easily modulated to adapt to different contexts, \ndedit{which stands in contrast to the goal of finding pitches that are universally deemed to be gender neutral, as previous work may suggest \cite{danielescu2020eschewing,mooshammer2021social}.} 

\noindent\textbf{\ndedit{Fine-grained control of voice is important for effective design modification.}} 
\ndedit{The voice design study revealed that changing one aspect of voice in a highly controlled manner allowed us to design voices with identities that reflect different genders, and, importantly, allowed us to not affect other aspects of the voice: realism and clarity (shown in \textbf{DP2} of the voice study) and also the perceived professionalism of the voice (shown in \textbf{DP3} of the voice study). We found in \textbf{H4a}, however, that} voice can be a more salient indicator of a robot's gender than its visual appearance, \ndedit{despite visual appearance setting gender in isolation (shown in \textbf{DP1} of the clothing design study). If designers aim to follow participatory design principles and align with ethical frameworks such as the Feminist HRI Framework \cite{winkle2023feminist} or the Robots for Social Justice Framework \cite{zhu2024robots}, tools as simple as modifying voice pitch can allow users to easily modify the perceived gender of the robot. }


\subsection{\ndedit{Clothing Design Implications}}

\noindent\textbf{\ndedit{Clothing design offers an effective way to modify existing robot embodiments.}}
\ndedit{This work was the first to incorporate a clothing design framework to modify a robot's gender.} We found that insights from fashion design on silhouette and style lines can be effectively used to change how participants perceive the gender of a robot, \ndedit{as shown in \textbf{DP1} of the clothing study}. Clothing provides a cost-effective and accessible alternative to changing the mechanical design of robots, allowing more diverse designers to have productive input in the creation of robots that interact with diverse populations of users. A key strength of this work is in the replication of our findings in two different contexts, indicating that considering silhouette and lines as attributes of design can be transferred between task contexts.

\ndedit{Like the voice design process, framing the clothing design process with specific elements of fashion allowed us to create precise modifications that affected the perceived gender of the robot, but did not affect other qualities such as perceived cost and quality (evaluated in \textbf{DP2} of the clothing design study). We also found that this framework was flexible enough to} reinforce expectation about what the task of the robot should be, \ndedit{as shown in \textbf{DP3} of the clothing study}. Using specific garments that are associated with tasks such as medical professionals and receptionists can effectively set the expectation for the robot's task. This is a key consideration in design, because when expectations align with reality, interaction outcomes are improved, and, conversely, they may decline if user expectations and reality are misaligned \cite{desai2012effects,cha2015perceived,khadpe2020conceptual}. 


\noindent\textbf{\ndedit{In context, data-driven evaluations of perceptions of appearance are necessary.}}
Several works have selected robots to use in gender-evaluation studies because they are described as being androgynous, and perform manipulations with the assumption that the robot is perceived as androgynous \cite{jackson2019tact,jackson2020exploring,bryant2020should}. In this work, we similarly selected a robot that was designed to be androgynous. However, during the clothing design study, we found that users perceived the robot's embodiment as feminine, as shown in \textbf{DP1} of the clothing design study. We also found in evaluating \textbf{H4b} that clothing did not reliably modify gender when other gendered cues were introduced, despite \textbf{DP1} showing that in isolation, clothing did gender the robot. 
\ndedittwo{This has important implications for critically investigating power structures in robot deployments to determine how robots may be reinforcing or upholding gender-based stereotypes.} 
Engaging in the analysis of power structures purely based on the claims made by designers can result in flawed interpretations of the role the robot is actually fulfilling. By using the evaluation techniques we outlined in our studies, designers may more effectively determine societal perceptions of robots. By knowing these social perceptions,  designers can better assess how these robots are situated in societal power structures.



\subsection{\ndedit{Task Design Implications}}

\noindent\textbf{\ndedit{Social role in the task may supersede gender perception when socially evaluating robots.}}
In our video study, evaluating \textbf{H1} showed that there were significant social perception differences between the tasks. Participants more positively evaluated the robot with a lower social role \ndedit{even when this conflicts with task expectations. For example, in \textbf{H1} we found that the medical professional was rated as less competent than the hotel receptionist}. This indicates that participants were more comfortable when they perceived that the human interlocutor had some level of control in the interaction, consistent with related findings from other works \cite{li2015observer, reinhardt2017dominance}. \ndedit{However, in \textbf{H2} we found that there were no changes in social perception due to the robot's gender when performing the task. This suggests that gender may not affect interaction with a robot as strongly as other contextual factors of a task. Prior research that finds differences in robot gender and task may need to be reinterpreted through other ethical frameworks that examine contextual factors like power dynamics of a task rather than strictly the gender of the robot to better reconcile their findings. }


\subsection{\ndedit{Implications for Designing Gender in Robots}}

\noindent\textbf{\ndedit{Androgyny is not a panacea for stereotyping.}}
\ndedit{Previous works and responses to community surveys of HRI researchers have considered the idea that androgyny is a good design goal to reduce the stereotyping of robots \cite{pandey2018pepper,specian2021quori,wessel2023gender}. Our results show that this may in fact encourage users to stereotype the robot. We} found that social role, clothing, and voice have complex interactions in influencing how users perceive a robot's gender. \ndedit{In evaluating \textbf{H5}, we found that when the robot has an androgynous voice,} the perceived gender changed to reflect stereotypes of the task when the robot wore clothes that reflected the task, \ndedit{regardless of the gendered qualities of the clothes---}the perceived gender of the robot became more masculine and less feminine when the robot was engaged in a higher social role task, while the robot was viewed as less masculine when it was engaged in a lower social role task. This suggests that \ndedit{a robot engaged in a task may be interpreted as a performative act, and affect how the robot is gendered}. Previous work also found that androgynous robots performing gendered actions on their own affected how observers gendered the robot \cite{acskin2023gendered}, and our findings suggests that this extends to tasks where robots interact with people.

Gender perceptions vary based on other contextual factors as well, such as culture \cite{robertson2018robo}, making the design goal of developing universally androgynous robots ill-posed. If androgyny is a design criterion, gendered cues must be adapted for the specific task and context the robot is deployed in. \ndedit{We believe that combining our design methodologies with principles from ethical design frameworks can help address the problem of culturally contextual gender power structures.}

\noindent\textbf{\ndedit{Mixing gendered aesthetics should be further explored.}}
\ndedit{Previous work cautioned that mixing robot aesthetics resulted in} increased eeriness \cite{mitchell2011mismatch}, \ndedit{and this has been used to justify not mixing gendered aesthetics in robots \cite{paetzel2016congruency,galatolo2023right}. By creating robots to have only hyper-feminine or hyper-masculine identities, robot designers can reinforce a gender binary. This work found that mixing gendered aesthetics did not negatively impact social expectations. In evaluating \textbf{H3},} we found no signifcant difference in ratings of the robots' warmth, competence, and discomfort when we mixed traditionally feminine and masculine voices and clothing. This is consistent with related findings that children gave no differences in eeriness ratings of a robot with mixed feminine and masculine faces and voices \cite{paetzel2016effects}.

\ndedit{Mixing gendered cues} opens the design space of robots designed for particular tasks. \ndedit{Previous work has found that representation of underrepresented groups in robotics through robot identities can increase participation in robotics-related fields \cite{miranda2024case}.} Additionally, the subversion of gender expectations may serve to increase representation of marginalized genders in robots \cite{winkle2021boosting} while not having a large effect on the social expectations of the robot. Importantly, representing these identities may lead to increased participation in STEM-related fields \cite{winkle202315}, \ndedittwo{where women and queer identities are historically underrepresented \cite{queerinai2023queer,korpan2024launching,fields2024underrepresentation}}
Modifying robot identities through voice, appearance, and task may additionally mitigate the negative effects of marginalization in society at large by exposing the general public to diverse forms of social identities \cite{brewer2000reducing}.

\subsection{\ndedit{Limitations and Future Work}}

\ndedit{This work is limited in several ways. An overarching limitation of our work is that we conducted our studies online. This allowed us to reach a wider audience and more representatively sample the United States population. However,} in-person interaction with robots can suggest different gendered perceptions than online studies \cite{orefice2016let}. 

We were limited by using only a single robot embodiment. We focused on an anthropomorphic robot because such robots tend to be perceived as gendered. While anthropomorphism plays a large role in how gendered robots are perceived, other forms of robots (e.g., animal-like or object-like robots) also have the potential to be gendered \cite{dennler2022using}. 

\ndedit{We found that our recruited population skewed slightly toward white men compared to the general population of America. We believe that our design methodology can be readily replicated for evaluation with other populations. Future work may investigate how marginalized populations, especially the queer community, may have different views and conceptualizations of a robot's gender.} 

A limitation of \ndedit{the voice design study is that} we only examined a set of six voices, and only modulated pitch in the integrative video study. The space of voice design is much larger than just pitch \cite{cambre2019one}. \ndedit{Future studies could investigate what other factors of voice are important in establishing aspects of a robots identity, and how designers may modify voices to effectively set a robot's identity.}

A limitation of our robot clothing design is that we made intentionally subtle modulations to the clothes designed for the robot. While these clothes represented typical Western uniforms for the two tasks we studied and they elicited differences in perceived gender, future work may explore how more exaggerated displays of feminine and masculine aesthetics in clothing can affect perceptions of the robot as well as task alignment. This study also suggests supplementary work in different aesthetics of clothing beyond masculinity and femininity. \ndedit{Past work has shown that users have preferences based on the formality of clothing that shape their perceptions of the robot~\cite{ashok2023social}. Future work can explore how other styles} such as Victorian, grunge, or camp aesthetics, and how these may shape social perception of robots, especially in communities that \ndedit{form identities around these aesthetics.} Future work \ndedit{can extend our design methodologies to allow users} to personalize robot identities for specific tasks. Analogously, previous work has found individual differences in voice preference \cite{CHANG2018194} and appearance preferences \cite{fitter2021you}, and engaging in voice personalization processes has shown increased ratings of interaction quality \cite{shi2023evaluating}.

\subsection{\ndedit{Conclusion}}
This work outlined a design process and design principles for eliciting perceptions of gender in robots. We investigated both the individual and combined role of voice and appearance on robot gender perception. We found that the perception of a robot's gender can be modulated through the careful design and evaluation of both visual and auditory modalities. \ndedittwo{These results suggest that a robot's identity is \textit{performative}, and thus constructed through the choices of the robot's designers and the robot's decision-making processes.} This work offers new insights into the construction of robot gender and the influences of separate and interacting modalities of voice and appearance. The results from our three user studies aim to generate new research questions about normative expectations of robot gender and social roles.





\begin{appendices}

\section{Declarations}\label{sec:declarations}

\subsection*{Funding}
This work was funded by a National Science Foundation Graduate Research Fellowship awarded to Nathan Dennler, under \#DGE-1842487.

\subsection*{Competing Interests}
Prof. Maja Matari\'c is an editorial board member of the International Journal of Social Robotics. 

All authors certify that they have no affiliations with or involvement in any organization or entity with any financial interest or non-financial interest in the subject matter or materials discussed in this manuscript.

\subsection*{Data Availability and Ethics Approval}

Data for the reported studies will be made available to researchers upon reasonable request, in accordance with the IRB protocol associated with the studies (\#UP-18-00510). All participants provided informed consent to participate in this study.

\end{appendices}

\bibliography{references}

\end{document}